%% file: AAAI-DeasisK.9337.tex
\def\year{2020}\relax
\documentclass[letterpaper]{article} 
\usepackage{aaai20}  
\usepackage{times}  
\usepackage{helvet} 
\usepackage{courier}  
\usepackage[hyphens]{url}  
\usepackage{graphicx} 
\usepackage{graphicx}  
\usepackage{amsfonts}
\usepackage{amsmath}
\usepackage{amssymb}
\usepackage{amsthm}
\usepackage{bbm}
\usepackage{bm}
\usepackage{subfig}
\usepackage{mathtools}
\usepackage{cleveref}
\usepackage{physics}
\usepackage{afterpage}

\usepackage[]{appendix}
\usepackage{algorithm}
\usepackage{algpseudocode}

\usepackage[dvipsnames]{xcolor} 

\usepackage{lipsum}
\DeclareMathSymbol{\shortminus}{\mathbin}{AMSa}{"39}

\DeclareMathOperator*{\argmax}{arg\,max}

\newcommand{\Ex}{\mathbb{E}}
\newcommand{\R}{\mathbb{R}}

\renewcommand{\vec}[1]{\textbf{\textit{#1}}}
\newcommand{\val}{Q}
\newcommand{\valt}{\val}

\newcommand{\target}{Q}
\newcommand{\params}{\mathbf{w}}
\newcommand{\Features}{\bm{\Phi}}
\newcommand{\features}{\bm{\phi}}

\newcommand{\bellmanop}{T}
\newcommand{\smashednorm}[1]{\left\lVert \smash{#1} \right\rVert}

\newtheorem{definition}{Definition}

\newtheorem{theorem}{Theorem}
\newtheorem{proposition}{Proposition}

\newtheorem{lemma}{Lemma}

\newtheorem{assumption}{Assumption}

\newcommand{\citei}[1]{\citeauthor{#1} (\citeyear{#1})}

\title{Fixed-Horizon Temporal Difference Methods\\for Stable Reinforcement Learning}

\author{\Large \textbf{Kristopher De Asis,\textsuperscript{\rm 1} Alan Chan,\textsuperscript{\rm 1,3} Silviu Pitis,\textsuperscript{\rm 2} Richard S. Sutton,\textsuperscript{\rm 1} Daniel Graves\textsuperscript{\rm 3}}\\
\textsuperscript{\rm 1}University of Alberta, \textsuperscript{\rm 2}University of Toronto, \textsuperscript{\rm 3}Huawei Technologies Canada, Ltd.\\
\{kldeasis, achan4, rsutton\}@ualberta.ca, spitis@cs.toronto.edu, daniel.graves@huawei.com
}

\begin{document}

\maketitle

\begin{abstract}
We explore fixed-horizon temporal difference (TD) methods, reinforcement learning algorithms for a new kind of value function that predicts the sum of rewards over a \textit{fixed} number of future time steps. To learn the value function for horizon $h$, these algorithms bootstrap from the value function for horizon $h-1$, or some shorter horizon. Because no value function bootstraps from itself, fixed-horizon methods are immune to the stability problems that plague other off-policy TD methods using function approximation (also known as ``the deadly triad''). Although fixed-horizon methods require the storage of additional value functions, this gives the agent additional predictive power, while the added complexity can be substantially reduced via parallel updates, shared weights, and $n$-step bootstrapping. We show how to use fixed-horizon value functions to solve reinforcement learning problems competitively with methods such as Q-learning that learn conventional value functions. We also prove convergence of fixed-horizon temporal difference methods with linear and general function approximation. Taken together, our results establish fixed-horizon TD methods as a viable new way of avoiding the stability problems of the deadly triad.
\end{abstract}

\section{Temporal Difference Learning}
\label{sec:tdintro}



\textit{Temporal difference} (TD) methods~\cite{sutton1988} are an important approach to reinforcement learning (RL) that combine ideas from Monte Carlo estimation and dynamic programming. A key view of TD learning is that it incrementally learns testable, predictive knowledge of the environment~\cite{horde2011}. The learned values represent answers to questions about how a signal will accumulate over time, conditioned on a way of behaving. 
In control tasks, this signal is the reward sequence, and the values represent an arbitrarily long sum of rewards an agent expects to receive when acting greedily with respect to its current predictions.

A TD learning agent's prediction horizon is specified through a \textit{discount factor}~\cite{rlbook2018}. This parameter adjusts how quickly to exponentially decay the weight given to later outcomes in a sequence's sum, and allows computational complexity to be \textit{independent of span}~\cite{indspan2015}. It's often set to a constant $\gamma \in [0, 1)$, but prior work generalizes the discount rate to be a transition-dependent \textit{termination} function~\cite{taskspec2016}. This allows for (variable) finite-length sums dependent on state transitions, like in episodic tasks.

In this paper, we explore a case of \textit{time-dependent} discounting, where the sum considers a fixed number of future steps regardless of where the agent ends up. We derive and investigate properties of fixed-horizon TD algorithms, and identify benefits over infinite-horizon algorithms in both prediction and control. 
%
Specifically, by storing and updating a separate value function for each horizon, fixed-horizon methods avoid feedback loops when bootstrapping, so that learning is stable even in presence of function approximation. Fixed-horizon agents can approximate infinite-horizon returns arbitrarily well, expand their set of learned horizons freely (computation permitting), and combine forecasts from multiple horizons to make time-sensitive predictions about rewards.
We emphasize 
our novel focus on predicting a fixed-horizon return from each state as a solution method, regardless of the problem setting. Our algorithms can be applied to both finite-horizon and infinite-horizon MDPs.

\section{MDPs and One-step TD Methods}\label{sec:mdps}

The RL problem is usually modeled as a Markov decision process (MDP), in which an agent interacts with an environment over a sequence of discrete time steps. 

At each time step $t$, the agent receives information about the environment's current \textit{state}, $S_t \in \mathcal{S}$, where $\mathcal{S}$ is the set of all possible states in the MDP. The agent uses this state information to select an \textit{action}, $A_t \in \mathcal{A}(S_t)$, where $\mathcal{A}(s)$ is the set of possible actions in state $s$. Based on the current state and the selected action, the agent gets information about the environment's next state, $S_{t+1} \in \mathcal{S}$, and receives a \textit{reward}, $R_{t+1} \in \mathbb{R}$, according to the \textit{environment model}, $p(s',r|s,a)={P(S_{t+1}=s',R_{t+1}=r|S_t=s,A_t=a)}$.

Actions are selected according to a \textit{policy}, $\pi(a|s) = P(A_t=a|S_t=s)$, which gives the probability of taking action $a$ given state $s$. An agent is interested in the return:
\begin{equation}
G_t = \sum_{k=0}^{T-t - 1} \gamma^{k} R_{t+k+1}
\label{eqn:returndef}
\end{equation}
where $\gamma \in [0,1]$ and $T$ is the final time step in an episodic task, and $\gamma \in [0,1)$ and $T = \infty$ for a continuing task.

\textit{Value-based methods} approach the RL problem by computing \textit{value functions}. In prediction, or \textit{policy evaluation}, the goal is to accurately estimate a policy's expected return, and a \textit{state-value function}, denoted $v_\pi(s)$, is estimated. In control, the goal is to learn a policy that maximizes the expected return, and an \textit{action-value function}, denoted $q_\pi(s, a)$, is estimated. In each case, the value functions represent a policy's expected return from state $s$ (and action $a$):
\begin{align}
v_\pi(s) &= \mathbb{E}_\pi[G_t|S_t=s]
\label{eqn:vdef}\\
q_\pi(s,a) &= \mathbb{E}_\pi[G_t|S_t=s,A_t=a]
\label{eqn:qdef}
\end{align}

TD methods learn to approximate value functions by expressing Equations \ref{eqn:vdef} and \ref{eqn:qdef} in terms of successor values (the \textit{Bellman equations}). The Bellman equation for $v_\pi$ is:
\begin{equation}
v_\pi(s) = \sum_{a}{\pi(a|s)\sum_{s',r}{p(s',r|s,a)\Big(r + \gamma v_\pi(s')\Big)}}
\label{eqn:vbellman}
\end{equation}
Based on Equation \ref{eqn:vbellman}, one-step TD prediction estimates the return by taking an action in the environment according to a policy, sampling the immediate reward, and bootstrapping off of the current estimated value of the next state for the remainder of the return. 
The difference between this \textit{TD target} and the value of the previous state (the \textit{TD error}) is then computed, and the previous state's value is updated by taking a step proportional to the TD error with $\alpha \in (0, 1]$:
\begin{align}
\hat{G}_t &= R_{t+1} + \gamma V(S_{t+1})
\label{eqn:tdtarget} \\
V(S_t) &\leftarrow V(S_t) + \alpha [\hat{G}_t - V(S_t)]
\label{eqn:tdupdate}
\end{align}
Q-learning~\cite{watkinsq1989} is arguably the most popular TD method for control. It has a similar update, but because policy improvement is involved, its target is a sample of the Bellman optimality equation for action-values:
\begin{align}
\hat{G}_t &= R_{t+1} + \gamma \max_{a'}{Q(S_{t+1}, a')}
\label{eqn:qltarget} \\
Q(S_t, A_t) &\leftarrow Q(S_t, A_t) + \alpha [\hat{G}_t - Q(S_t, A_t)]
\label{eqn:qlupdate}
\end{align}


For small finite state spaces where the value function can be stored as a single parameter vector, also known as the \textit{tabular} case, TD methods 
are known to converge under mild technical conditions 
\cite{rlbook2018}. For large or uncountable state spaces, however, one must use function approximation 
to represent the value function, which does not have the same convergence guarantees. 

Some early examples of divergence with function approximation were provided by \citei{boyan1995generalization}, who proposed the Grow-Support algorithm to combat divergence. \citei{baird1995} provided perhaps the most famous example of divergence (discussed below) and proposed residual algorithms. \citei{gordon1995stable} proved convergence for a specific class of function approximators known as \textit{averagers}. Convergence of TD prediction using linear function approximation, first proved by \citei{tsitsiklis1997analysis}, requires the training distribution to be on-policy. This approach was later extended to Q-learning \cite{melo2007q}, but under relatively stringent conditions. In particular, Assumption (7) of Theorem 1 in ~\citeauthor{melo2007q}~\shortcite{melo2007q} amounts to a requirement that the behaviour policy is already somewhat close to the optimal policy. 

\newcommand{\triad}[1]{\,\raisebox{-0.26\baselineskip}{$^{(#1)}$}}
The on-policy limitations of the latter two results reflect what has come to be known as the \textit{deadly triad} \cite{rlbook2018}: when using \triad{1} TD methods with \triad{2} function approximation, training on \triad{3} an off-policy data distribution can result in instability and divergence. One response to this problem is to shift the optimization target, as done by Gradient TD (GTD) methods \cite{sutton2009fast,bhatnagar2009convergent}; while provably convergent, GTD algorithms are empirically slow \cite{ghiassian2018online,rlbook2018}. Another approach is to approximate Fitted Value Iteration (FVI) methods \cite{munos2008finite} using a \textit{target network}, as proposed by \citei{dqn2015}. Though lacking convergence guarantees, this approach has been empirically successful. In the next section, we propose fixed-horizon TD methods, an alternative approach to ensuring stability in presence of the deadly triad.

\section{Fixed-horizon TD Methods}\label{sec:fhreturns}


A \textit{fixed-horizon return} is a sum of rewards similar to Equation \ref{eqn:returndef} that includes only a fixed number of future steps. 
For a fixed horizon $h$, the fixed-horizon return is defined to be:
\begin{equation}
G^h_t = \sum_{k=0}^{\min(h, T-t) - 1} \gamma^k R_{t+k+1}
\label{eqn:deffhreturn}
\end{equation}
which is well-defined for any finite $\gamma$. 
This formulation allows the agent's horizon of interest and sense of urgency to be characterized more flexibly and admits the use of $\gamma = 1$ in the continuing setting. 
Fixed-horizon value functions are defined as expectations of the above sum when 
following policy $\pi$ beginning with state $s$ (and action $a$):
\begin{align}
v^h_\pi(s) &= \mathbb{E}_\pi[G^h_t|S_t=s]
\label{eqn:deffhstatevalues} \\
q^h_\pi(s, a) &= \mathbb{E}_\pi[G^h_t|S_t=s, A_t=a]
\label{eqn:deffhactionvalues}
\end{align}
These fixed-horizon value functions can also be written in terms of successor values. Instead of bootstrapping off of the same value function, $v^h_\pi(s)$ bootstraps off of the successor state's value from an earlier horizon~\cite{bertsekastwo2012}:
\begin{equation}\small
v^h_\pi(s) = \sum_{a}{\pi(a|s)\sum_{s', r}{p(s',r|s,a)\Big(r + \gamma v^{h-1}_\pi(s')\Big)}}
\label{eqn:fhvbellman}
\end{equation}
where $v^{0}_{\pi}(s) = 0$ for all $s \in \mathcal{S}$.

From the perspective of \textit{generalized value functions} (GVFs)~\cite{horde2011}, fixed-horizon value functions are compositional GVFs. 
If an agent knows about a signal up to a final horizon of $H$, it can specify a question under the same policy with a horizon of $H + 1$, and directly make use of existing values. 
As $H \to \infty$, the fixed-horizon return converges to the infinite-horizon return, so that a fixed-horizon agent with sufficiently large $H$ could solve infinite-horizon tasks arbitrarily well. Indeed, for an infinite-horizon MDP with absolute rewards bounded by $R_{max}$ and $\gamma < 1$, it is easy to see that $|v_\pi^h(s) - v_\pi(s)| \leq \gamma^h\frac{R_{max}}{1-\gamma}$.

For a final horizon $H \ll \infty$, there may be concerns about suboptimal control. We explore this empirically in Section \ref{sec:empirical}. For now, we note that optimality is never guaranteed when values are approximated. This can result from function approximation, or even in tabular settings from the use of a constant step size. Further, recent work shows benefits in considering shorter horizons~\cite{lowergamma2019} based on performance metric mismatches.

\subsection{One-step Fixed-horizon TD}

One-step fixed-horizon TD (FHTD) learns approximate values $V^h \approx v^h_\pi$ by computing, for each $h \in \{1, 2, 3, ..., H\}$:
\begin{align}
\hat{G}^h_t &= R_{t+1} + \gamma V^{h-1}(S_{t+1})
\label{eqn:onestepfhtdtarget} \\
V^h(S_t) &\leftarrow V^h(S_t) + \alpha [\hat{G}^h_t - V^h(S_t)]
\label{eqn:onestepfhtdupdate}
\end{align}
where $V^0(s) = 0$ for all $s \in \mathcal{S}$.
The general procedure of one-step FHTD was previously described by~\citei{sutton1988}, but not tested due to the observation that one-step FHTD's computation and storage scale linearly with the final horizon $H$, as it performs $H$ value updates per step. We argue that because these value updates can be parallelized, reasonably long horizons are feasible, and we'll later show that $n$-step FHTD can make even longer horizons practical. Forms of temporal abstraction~\cite{options1999} can further substantially reduce the complexity.

A key property of FHTD is that bootstrapping is grounded by the 0th horizon, which is \textit{exactly} known from the start. The 1st horizon's values estimate the expected immediate reward from each state, which is a stable target. The 2nd horizon bootstraps off of the 1st, which eventually becomes a stable target, and so forth. This argument intuitively applies even with general function approximation, assuming weights are not shared between horizons. To see the implications of this on TD's deadly triad, consider one-step FHTD with linear function approximation. For each horizon's weight vector, $\params^h$, the following update is performed:
\begin{equation*}
    \params^h_{t+1} \leftarrow \params^h_{t} + \alpha \big[R_{t+1} + \gamma \params^{h-1}_{t} \cdot \features_{t+1} - \params^h_t \cdot \features_t \big] \features_t
\end{equation*}
where $\features_t$ is the feature vector of the state at time $t$, $\features(S_t)$. The expected update can be written as:
\begin{equation*}
    \params^h_{t+1} \leftarrow \params^h_t + \alpha [\mathbf{b} - \mathbf{A} \params^h_t]
\end{equation*}
where $\mathbf{A} = \mathbb{E}[\features_t\features_t^T]$ and $\mathbf{b} = \mathbb{E}[(R_{t+1} + \gamma  \params^{h-1}_t \cdot \features_{t+1}) \features_t]$, having expectations over the transition dynamics and the sampling distribution over states. 
Because the target uses a different set of weights, the $\mathbf{b}$ vector is non-stationary (but convergent), and the $\mathbf{A}$ matrix is an expectation over the \textit{fixed} sampling distribution. 
Defining $\Features$ to be the matrix containing each state's feature vector along each row, and $\mathbf{D}$ to be the diagonal matrix containing the probability of sampling each state on its diagonal, we have that $\mathbf{A} = \Features^T \mathbf{D} \Features$. $\mathbf{A}$ is always positive definite and so the updates are guaranteed to be stable relative to the current $\mathbf{b}$ vector. In contrast, TD's update gives $\mathbf{A} = \Features^T \mathbf{D} (I - \gamma \mathbf{P}) \Features$ where $\mathbf{P}$ contains the state-transition probabilities under the target policy.
TD's $\mathbf{A}$ matrix can be negative definite if $\mathbf{D}$ and $\mathbf{P}$ do not match (i.e., off-policy updates), and the weights may diverge. We explore the convergence of FHTD formally in Section \ref{sec:convergence}.

A horizon's reliance on earlier horizons' estimates being accurate may be expected to have worse sample complexity. If each horizon's parameters are identically initialized, distant horizons will match infinite-horizon TD's updates for the first $H - 1$ steps. If $H\rightarrow\infty$, this suggests that FHTD's sample complexity might be upper bounded by that of TD. See Appendix \ref{appdx:samplecomplexity} for a preliminary result in this vein.

An FHTD agent has strictly more predictive capabilities than a standard TD agent. FHTD can be viewed as computing an inner product between the rewards and a step function. This gives an agent an exact notion of \textit{when} rewards occur, as one can subtract the fixed-horizon returns of subsequent horizons to get an individual expected reward at a specific future time step. 

\subsection{Multi-step Fixed-horizon TD Prediction}
\label{sec:nstepfhtd}

Another way to estimate fixed-horizon values is through \textit{Monte Carlo} (MC) methods. From a state, one can generate reward sequences of fixed lengths, and average their sums into the state's expected fixed-horizon return. Fixed-horizon MC is an opposite extreme from one-step FHTD in terms of a bias-variance trade-off, similar to the infinite-horizon setting~\cite{multistepbiasvar1994}. This trade-off motivates considering multi-step FHTD methods.

Fixed-horizon MC is appealing when one only needs the expected return for the final horizon $H$, and has no explicit use for the horizons leading up to $H$. This is because it can learn the final horizon directly by storing and summing the last $H$ rewards, and performing one value-function update for each visited state. If we use a fixed-horizon analogue to $n$-step TD methods \cite{rlbook2018}, denoted $n$-step FHTD, the algorithm stores and sums the last $n$ rewards, and only has to learn $\lceil\frac{H}{n}\rceil$ value functions. For each $h \in \{n, 2n, 3n, ..., H\}$, $n$-step FHTD computes:
\begin{align}\small
\hat{G}^h_{t:t+n} &= \gamma^n V^{h - n}(S_{t+n}) + \sum^{n-1}_{k=0}{\gamma^k R_{t+k+1}}
\label{eqn:nstepfhtdtarget} \\
V^h(S_t) &\leftarrow V^h(S_t) + \alpha [\hat{G}^h_{t:t+n} - V^h(S_t)]
\label{eqn:nstepfhtdupdate}
\end{align}
assuming that $H$ is divisible by $n$. If $H$ is not divisible by $n$, the earliest horizon's update will only sum the first $H\ (\mathrm{mod}\ n)$ of the last $n$ rewards. Counting the number of rewards to sum, and the number of value function updates, $n$-step FHTD performs $n - 1 + H/n$ operations per time step. This has a worst case of $H$ operations at $n = 1$ and $n = H$, and a best case of $2\sqrt{H} - 1$ operations at $n = \sqrt{H}$. This suggests that in addition to trading off reliance on sampled information versus current estimates, $n$-step FHTD's computation can scale sub-linearly in $H$.

\subsection{Fixed-horizon TD Control}

The above FHTD methods for state-values can be trivially extended to learn action-values under fixed policies, but it is less trivial with policy improvement involved.

If we consider the best an agent can do from a state in terms of the next $H$ steps, it consists of an immediate reward, and the best it can do in the next $H - 1$ steps from the next state. That is, each horizon has a separate target policy that's greedy with respect to its own horizon. Because the greedy action may differ between one horizon and another, FHTD control is inherently off-policy.

Q-learning provides a natural way to handle this off-policyness, where the TD target bootstraps off of an estimate under a greedy policy. Based on this, fixed-horizon Q-learning (FHQ-learning) performs the following updates:
\begin{align}
\hat{G}^h_t &= R_{t+1} + \gamma \max_{a'}{Q^{h-1}(S_{t+1}, a')}
\label{eqn:onestepfhqltarget} \\
Q^h(S_t, A_t) &\leftarrow Q^h(S_t, A_t) + \alpha [\hat{G}^h_t - Q^h(S_t, A_t)]
\label{eqn:onestepfhqlupdate}
\end{align}

A saddening observation from FHTD control being inherently off-policy is that the computational savings of $n$-step FHTD methods may not be possible without approximations.
With policy improvement, an agent needs to know the current greedy action for each horizon. This information isn't available if $n$-step FHTD methods avoid learning intermediate horizons.
On the other hand, a benefit of each horizon having its own greedy policy is that the optimal policy for a horizon is unaffected by the policy improvement steps of later horizons. As a compositional GVF, the unidirectional decoupling of greedy policies suggests that in addition to a newly specified final horizon leveraging previously learned values for prediction, it can leverage \textit{previously learned policies} for control.

\section{Convergence of Fixed-horizon TD}\label{sec:convergence}

This section sets forth our convergence results, first for linear function approximation (which includes the tabular case), and second for general function approximation.

\subsection{Linear Function Approximation}
For linear function approximation, we prove the convergence of FHQ-learning under some additional assumptions, outlined in detail in Appendix \ref{appdx:convergence}. Analogous proofs may be obtained easily for policy evaluation. We provide a sketch of the proof below, and generally follow the outline of ~\citeauthor{melo2007q}~\shortcite{melo2007q}. Full proofs are in the Appendix.

We denote $\phi(s, a) \in \R^d$ for the feature vector corresponding to the state $s$ and action $a$. We will sometimes write $\phi_s$ for $\phi(s, a)$. Assume furthermore that the features are linearly independent and bounded. 




We assume that we are learning $H$ horizons, and approximate the fixed-horizon $h$-th action-value function linearly.
\begin{equation*}
    Q^h_\params(s, a) :=  \params^{h, :} \phi(s, a),
\end{equation*}

\noindent where $\params \in \R^{H \times d}$. Colon indices denote array slicing, with the same conventions as in NumPy. For convenience, we also define $\params^{0, :} := 0$. 

We define the feature corresponding to the max action of a given horizon:
\begin{equation*}
    \phi_{\params}^*(s, h) := \phi(s, \argmax_{a} Q^h_\params(s, a)).
\end{equation*}

\noindent The update at time $t + 1$ for each horizon $h$ is given by 
\begin{align*}
    \params_{t + 1}^{h + 1, :} &:= \params_t^{h + 1, :} + \alpha_t (r(s, a, s') + \gamma \params_t^{h, :} \phi^*_{\params_t}(s', h) \\
    &\quad - \params_t^{h + 1, :} \phi_s)\phi_s^T.
\end{align*}
\noindent where the step-sizes $\alpha_t$ are assumed to satisfy:
$$
    \sum_{t} \alpha_t = \infty\quad\text{  and  }\quad
    \sum_{t} \alpha_t^2 < \infty.
$$

\begin{proposition}\label{prop:ode}



For $h = 1, ..., H$, the following ODE system has an equilibrium.
\begin{equation}\label{eq:ode}
    \dot{\params}^{h + 1, :} = \Ex\left[(r(s, a, s') + \gamma \params^{h, :} \phi^*_{\params}(s', h) - \params^{h + 1, :} \phi_s)\phi_s^T\right]
\end{equation}

\noindent Denote one such equilibrium by $\params_e$, and define $\tilde{\params} := \params - \params_e$. If, furthermore, we have that for $h = 1, ..., H$,
\begin{align}\label{eq:ode-bound}
    \gamma^2 \Ex[(\params^{h, :} \phi^*_{\params}&(s, h) - \params_e^{h, :} \phi^*_{\params_e}(s, h))^2 ] <\\
    &\quad \Ex\left[(\params^{h + 1, :} \phi_s - \params_e^{h + 1, :} \phi_s)^2\right]\nonumber 
\end{align}
\noindent then $\params_e$ is a globally asymptotically stable equilibrium of \Cref{eq:ode}. 

\end{proposition}

\begin{proof}
    See Appendix \ref{appdx:convergence}. The main idea is to explicitly construct an equilibrium of \Cref{eq:ode} and to use a Lyapunov function along with \Cref{eq:ode-bound} to show global asymptotic stability. 
\end{proof}

\noindent \Cref{eq:ode-bound} means that the $h$-th fixed-horizon action-value function must be closer, when taking respective max actions, to its equilibrium point than the $(h + 1)$-th fixed-horizon action-value function is to its equilibrium, where distance is measured in terms of squared error of the action-values. Intuitively, the functions for the previous horizons must have converged somewhat for the next horizon to converge, formalizing the cascading effect described earlier. This assumption is reasonable given that value functions for smaller horizons have a bootstrap target that is closer to a Monte Carlo target than the value functions for larger horizons. As a result, \cref{eq:ode-bound} is somewhat less stringent than the corresponding assumption (7) in Theorem 1 of ~\citeauthor{melo2007q}~\shortcite{melo2007q}, which requires the behaviour policy already be somewhat close to optimal. 

\begin{theorem}\label{thm:linearapproximation}
Viewing the right-hand side of the ODE system \cref{eq:ode} as a single function $g(\params)$, assume that $g$ is locally Lipschitz in $\params$.
Assuming also \Cref{eq:ode-bound}, a fixed behaviour policy $\pi$, and the assumptions in the Appendix, the iterates of FHQ-learning converge with probability 1 to the equilibrium of the ODE system \cref{eq:ode}.
\end{theorem}
\begin{proof}
See Appendix \ref{appdx:convergence}. The main idea is to use Theorem 17 of ~\citeauthor{benveniste1990stochastic}~\shortcite{benveniste1990stochastic} and \Cref{prop:ode}.
\end{proof}

A limitation of \Cref{thm:linearapproximation} is the assumption of a fixed behaviour policy. It is possible to make a similar claim for a changing policy, assuming that it satisfies a form of Lipschitz continuity with respect to the parameters. See~\citeauthor{melo2007q}~\shortcite{melo2007q} for discussion on this point. 

\subsection{General Function Approximation}
We now address the case where $\val^h$ is represented by a general function approximator (e.g., a neural network). As before, the analysis extends easily to prediction ($V^h$). General non-linear function approximators have non-convex loss surfaces and may have many saddle points and local minima. As a result, typical convergence results (e.g., for gradient descent) are not useful without some additional assumption about the approximation error (cf., the \textit{inherent Bellman error} in the analysis of Fitted Value Iteration \cite{munos2008finite})
. We therefore state our result for general function approximators in terms of \textit{$\delta$-strongness} for $\delta > 0$:

\begin{definition}\label{defition_delta_strongness}
	A function approximator, consisting of function class $\mathcal{H}$ and iterative learning algorithm $\mathcal{A}$, is \textbf{$\bm{\delta}$-strong} with respect to target function class $\mathcal{G}$ and loss function $J: \mathcal{H} \times \mathcal{G} \to \R^+$ if, for all target functions $g \in \mathcal{G}$, the learning algorithm $\mathcal{A}$ is \textit{guaranteed} to produce (within a finite $t$ number of steps) an $h_t \in \mathcal{H}$ such that $J(h_t, g) \leq \delta$. 
\end{definition}

We consider learning algorithms $\mathcal{A}$ that converge once some minimum progress $c$ can no longer be made:

\begin{assumption}\label{assumption_minimum_progress}
	There exists stopping constant $c$ such that algorithm $\mathcal{A}$ is considered ``converged'' with respect to target function $g$ if less than $c$ progress would be made by an additional step; i.e., if $J(h_{t}, g) - J(h_{t+1}, g) < c$. 
\end{assumption} 

Note that $\delta$-strongness may depend on stopping constant $c$: a larger $c$ naturally corresponds to earlier stopping and looser $\delta$. Note also that, so long as the distance between the function classes $\mathcal{H}$ and $\mathcal{G}$ is upper bounded, say by $\vec{d}$, any convergent $\mathcal{A}$ is ``$\vec{d}$-strong''. Thus, a $\delta$-strongness result is only meaningful to the extent that $\delta$ is sufficiently small. 

We consider functions $\valt^h_t = \valt(\params^{h}_t)$ parameterized by $\params^{h}_t$. Letting $\bellmanop$ denote the Bellman operator $[\bellmanop\valt](s, a) = \mathbb{E}_{s',r \sim p(\cdot|s,a)}[r + \gamma \max_{a'}\valt(s',a')]$, we assume:

\begin{assumption}
	\label{assumption_Lipschitz}
		The target function $\bellmanop\valt(\params)$ is Lipschitz continuous in the parameters: there exists constant $L$ such that $\smashednorm{\bellmanop\valt(\params_1) - \bellmanop\valt(\params_2)}_\mathcal{F}\leq L\smashednorm{\params_1 - \params_2}$ for all $\params_1, \params_2$, where $\smashednorm{\cdot}_\mathcal{F}$ is a norm on value function space $\mathcal{F}$ (typically a weighted $L_2$ norm, weighted by the data distribution), which we take to be a Banach space containing both $\mathcal{H}$ and $\mathcal{G}$.
\end{assumption}

It follows that if $(\params^{h-1}_t) \to \params^{h-1}_*$, the sequence of target functions $\bellmanop\valt^{h-1}_t$ converges to $\bellmanop\target^{h-1}_* = \bellmanop\valt(\params^{h-1}_*)$ in $\mathcal{F}$ under norm $\smashednorm{\cdot}$. We can therefore define the ``true'' loss:
\begin{equation}
	J^*(\params^h_t) = \smashednorm{\bellmanop\target^{h-1}_* - \valt^h_t}
\end{equation}

\noindent where we drop the square from the usual mean square Bellman error \cite{rlbook2018} for ease of exposition (the analysis is unaffected after an appropriate adjustment). Since we cannot access $J^*$, we optimize the surrogate loss:
\begin{equation}
	J(\params^{h-1}_t, \params^h_t) = \smashednorm{\bellmanop\valt^{h-1}_t - \valt^h_t}.
\end{equation}
\begin{lemma}\label{lemma_strict_progress}
	If $\smashednorm{\bellmanop\valt^{h-1}_t - \bellmanop\target^{h-1}_*} < \epsilon$, 
	and learning has not yet converged with respect to the surrogate loss $J$, then $J^*(\params^h_t) -J^*(\params^h_{t+1}) > c - 2\epsilon$.
\end{lemma}

\newcommand{\moveback}{}
\begin{proof}
	Intuitively, enough progress toward a similar enough surrogate loss guarantees progress toward the true loss. 
	Applying the triangle inequality (twice) gives:
	\begin{equation*}\small
    		J^*(\params^h_t) - J^*(\params^h_{t+1}) = \smashednorm{\bellmanop\target^{h\shortminus1}_* - \valt^h_t} - \smashednorm{\bellmanop\target^{h\shortminus1}_* - \valt^h_{t+1}}
	\end{equation*}
	\begin{equation*}
    	\begin{split}\small
    		&\moveback= \smashednorm{\bellmanop\target^{h\shortminus1}_* - \valt^h_t} \\ 
    		&\moveback\quad\quad\quad + (\smashednorm{\bellmanop\valt^{h\shortminus1}_t - \bellmanop\target^{h\shortminus1}_*} -  \smashednorm{\bellmanop\valt^{h\shortminus1}_t - \bellmanop\target^{h\shortminus1}_*}) \\ 
    		&\moveback\quad\quad\quad
    		- \smashednorm{\bellmanop\target^{h\shortminus1}_* + (\bellmanop\valt^{h\shortminus1}_t - \bellmanop\valt^{h\shortminus1}_t) - \valt^h_{t+1}}\\
    		&\moveback\geq (\smashednorm{\bellmanop\valt^{h\shortminus1}_t - \valt^h_t} - \smashednorm{\bellmanop\valt^{h\shortminus1}_t - \bellmanop\target^{h\shortminus1}_*})\\
    		&\moveback\quad\quad\quad - (\smashednorm{\bellmanop\valt^{h\shortminus1}_t - \valt^h_{t+1}}+\smashednorm{\bellmanop\target^{h\shortminus1}_*- \bellmanop\valt^{h\shortminus1}_t })\\
    		&\moveback= (\smashednorm{\bellmanop\valt^{h\shortminus1}_t - \valt^h_t} -\smashednorm{\bellmanop\valt^{h\shortminus1}_t - \valt^h_{t+1} }\\
    		&\moveback\quad\quad\quad - 2\smashednorm{\bellmanop\target^{h\shortminus1}_* -  \bellmanop\valt^{h\shortminus1}_t }
		> c - 2\epsilon,\\
	\end{split}
	\end{equation*}
	where the final inequality uses $ \smashednorm{\bellmanop\valt^{h\shortminus1}_t - \valt^h_t} - \smashednorm{\bellmanop\valt^{h\shortminus1}_t - \valt^h_{t+1}} \geq c$ from 
	Assumption \ref{assumption_minimum_progress}.
\end{proof}

It follows from Lemma \ref{lemma_strict_progress} that when $\epsilon$ is small enough---$\epsilon < \frac{c}{2} - k$ for some constant $k$---either the true loss $J^*$ falls by at least $k$, or learning has converged with respect to the current target $\bellmanop\valt^{h-1}_t$. Since $J^*$ is non-negative (so cannot go to $-\infty$), it follows that the \textit{loss} converges to a $\delta$-strong solution: $J^*(\params^h_t) \to d$ with $d \leq \delta$. Since there are only a finite number of $k$-sized steps between the current loss at time $t$ and $0$ (i.e., only a finite number of opportunities for the learning algorithm to have ``not converged'' with respect to the surrogate $J$), the parameters $\params^h_t$ must also converge. 

Since $Q^0_t = 0$ is stationary, it follows by induction that:

\begin{theorem}\label{thm:generalapproximation}
Under Assumptions \ref{assumption_minimum_progress} and \ref{assumption_Lipschitz}, each horizon of FHQ-learning converges to a $\delta$-strong solution when using a $\delta$-strong function approximator.
\end{theorem}

In contrast to Theorem \ref{thm:linearapproximation}, which applies quite generally to linear function approximators, $\delta$-strongness and Assumption \ref{assumption_minimum_progress} limit the applicability of Theorem \ref{thm:generalapproximation} in two important ways.

First, since gradient-based learning may stall in a bad saddle point or local minimum, neural networks are not, in general, $\delta$-strong for small $\delta$. Nevertheless, repeat empirical experience shows that neural networks consistently find good solutions \cite{zhang2016understanding}, and a growing number of theoretical results suggest that almost all local optima are ``good'' \cite{pascanu2014saddle,choromanska2015loss,pennington2017geometry}. For this reason, we argue that $\delta$-strongness is reasonable, at least approximately.

Second, Assumption \ref{assumption_minimum_progress} is critical to the result: only if progress is ``large enough'' relative to the error in the surrogate target is learning guaranteed to make progress on $J^*$. Without a lower bound on progress---e.g., if the progress at each step is allowed to be less than $2\epsilon$ regardless of $\epsilon$---training might accumulate an error on the order of $\epsilon$ at every step. In pathological cases, the sum of such errors may diverge even if $\epsilon \to 0$. As stated, Assumption \ref{assumption_minimum_progress} does not reflect common practice: rather than  progress being measured at every step, it is typically measured over several, say $k$, steps. This is because training targets are noisy estimates of the expected Bellman operator and several steps are needed to accurately assess progress. Our analysis can be adapted to this more practical scenario by making use of a target network \cite{dqn2015} to freeze the targets for $k$ steps at a time. Then, considering each $k$ step window as a single step in the above discussion, Assumption \ref{assumption_minimum_progress} is fair.
This said, intuition suggests that pathological divergence when Assumption \ref{assumption_minimum_progress} is not satisfied is rare, and our experiments with Deep FHTD Control show that training can be stable even with shared weights and no target networks.

\section{Empirical Evaluation}\label{sec:empirical}

This section outlines several hypotheses concerning fixed-horizon TD methods, experiments aimed at testing them, and the results from each experiment. Pseudo-code, diagrams, more experimental details, and additional experiments can be found in the supplementary material.

\subsection{Stability in Baird's Counterexample}

We hypothesize that FHTD methods provide a stable way of bootstrapping, such that divergence will not occur under off-policy updating with function approximation. To test this, we used Baird's counterexample~\cite{baird1995}, a 7-state MDP where every state has two actions. One action results in a uniform random transition to one of the first 6 states, and the other action results in a deterministic transition to the 7th state. Rewards are always zero, and each state has a specific feature vector for use with linear function approximation. It was presented with a discount rate of $\gamma = 0.99$, and a target policy which always chooses to go to the 7th state.

In our experiment, we used one-step FHTD with importance sampling corrections~\cite{montecarlo1981} to predict up to a horizon of $H = \frac{1}{1-\gamma}=100$. Each horizon's weights were initialized to be $\textbf{w}^h = [1, 1, 1, 1, 1, 1, 10, 1]^T$, based on~\citeauthor{rlbook2018}~\shortcite{rlbook2018}, and we used a step size of $\alpha = \frac{0.2}{ \lvert \mathcal{S} \rvert}$. We performed 1000 independent runs of 10,000 steps, and the results can be found in Figure \ref{fig:results_baird}.

\newcommand{\figwidth}{7cm}
\begin{figure}[!bp]
    \centering
	\includegraphics[width=\figwidth]{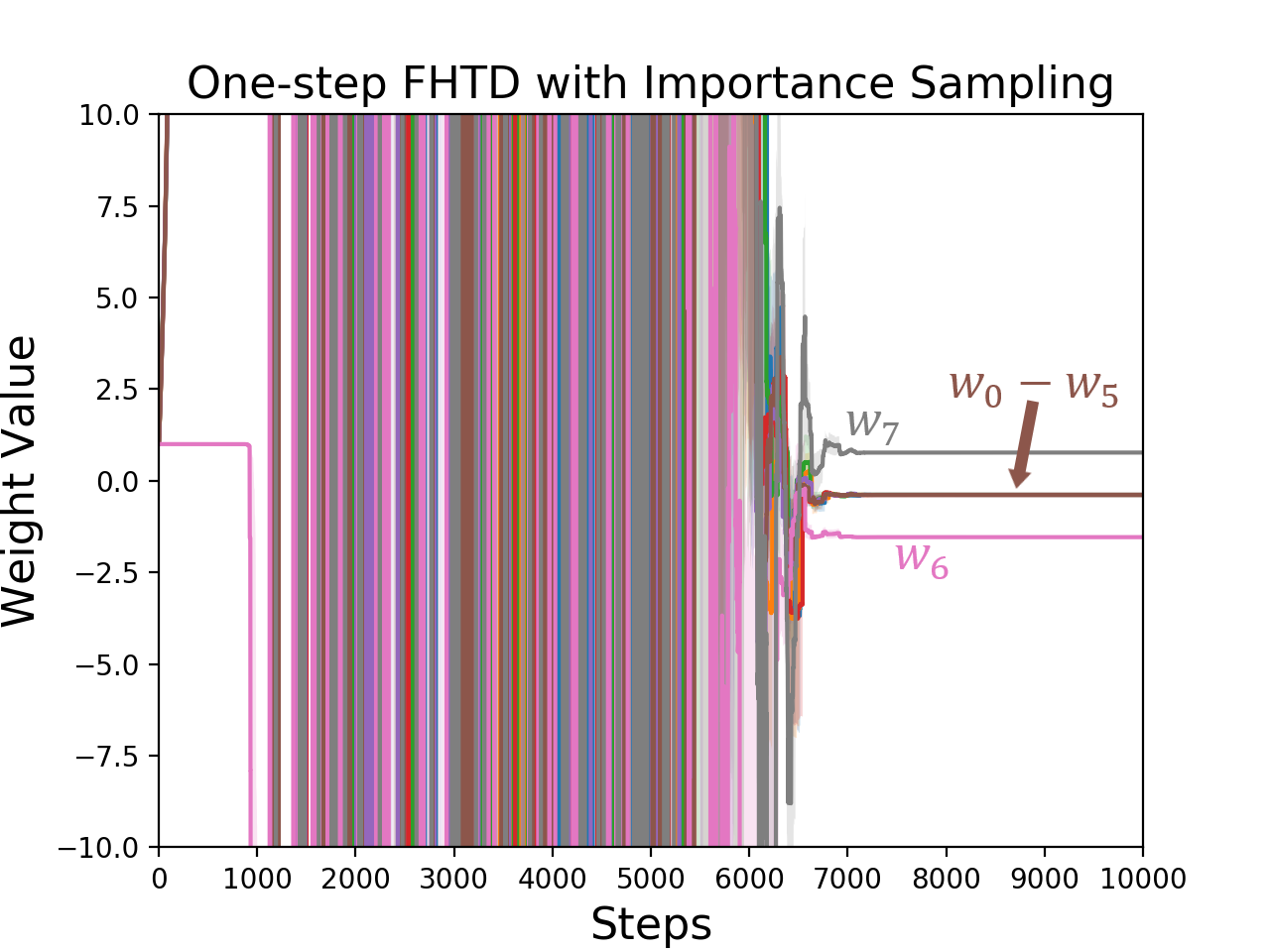}
	\caption[]{Weight trajectories of one-step FHTD's 100th horizon value function on Baird's counterexample, plotted after each time step. Shaded regions represent one standard error.}
	\label{fig:results_baird}
\end{figure}

We see that one-step FHTD eventually and consistently converges. The initial apparent instability is due to each horizon being initialized to the same weight vector, making early updates resemble the infinite-horizon setting where weight updates bootstrap off of the same weight vector. The results emphasize what TD would have done, and how FHTD can recover from it. Of note, the final weights do give optimal state-values of $0$ for each state. In results not shown, FHTD still converges, sometimes quicker, when each horizon's weights are initialized randomly (and not identically). 

\subsection{Tabular FHTD Control}\label{subsec:tabularFHTDcontrol}
\begin{figure*}[!btp]
    \centering
	\includegraphics[width=\figwidth]{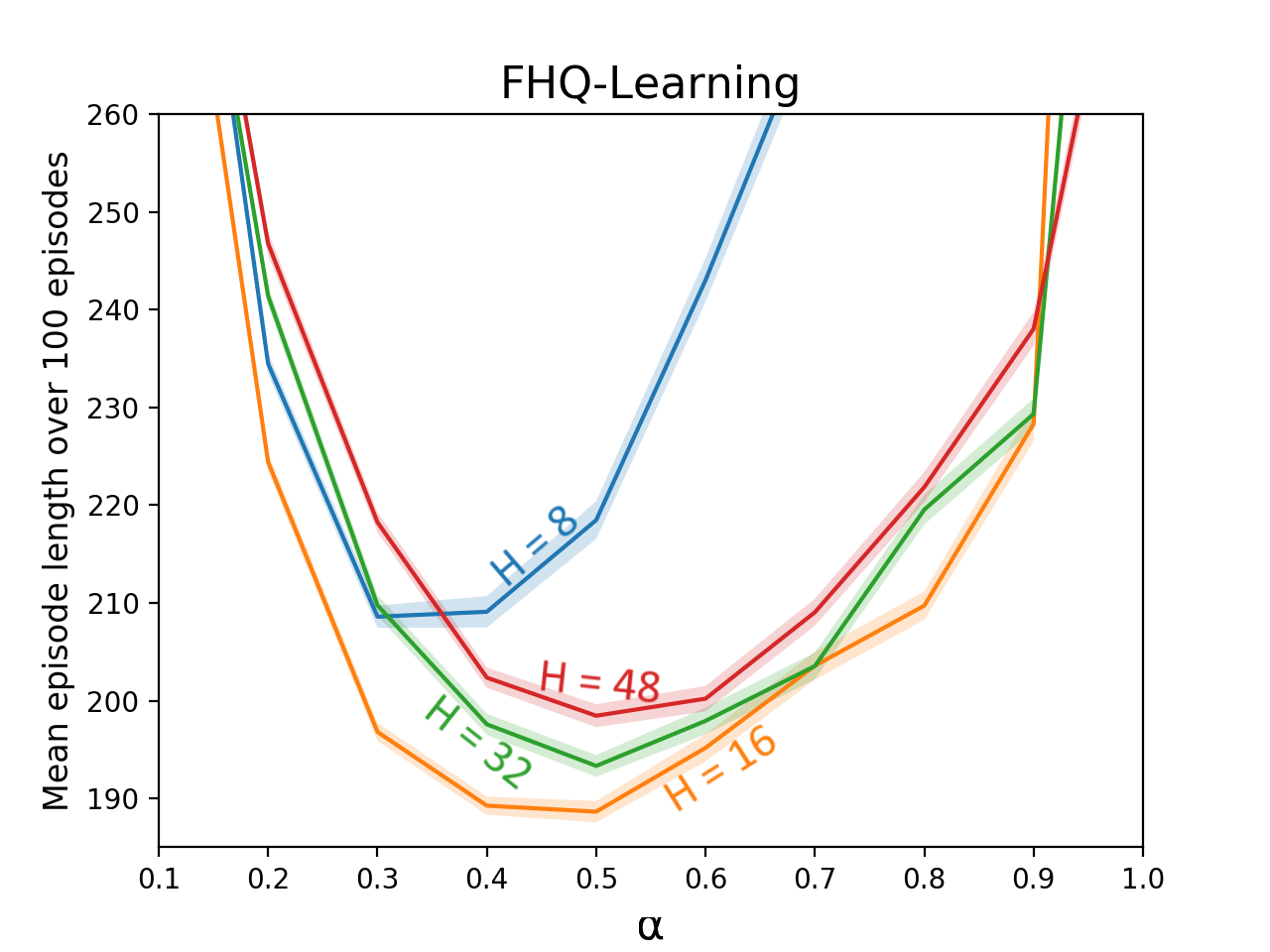}
	\hspace{0.05\linewidth}
	\includegraphics[width=\figwidth]{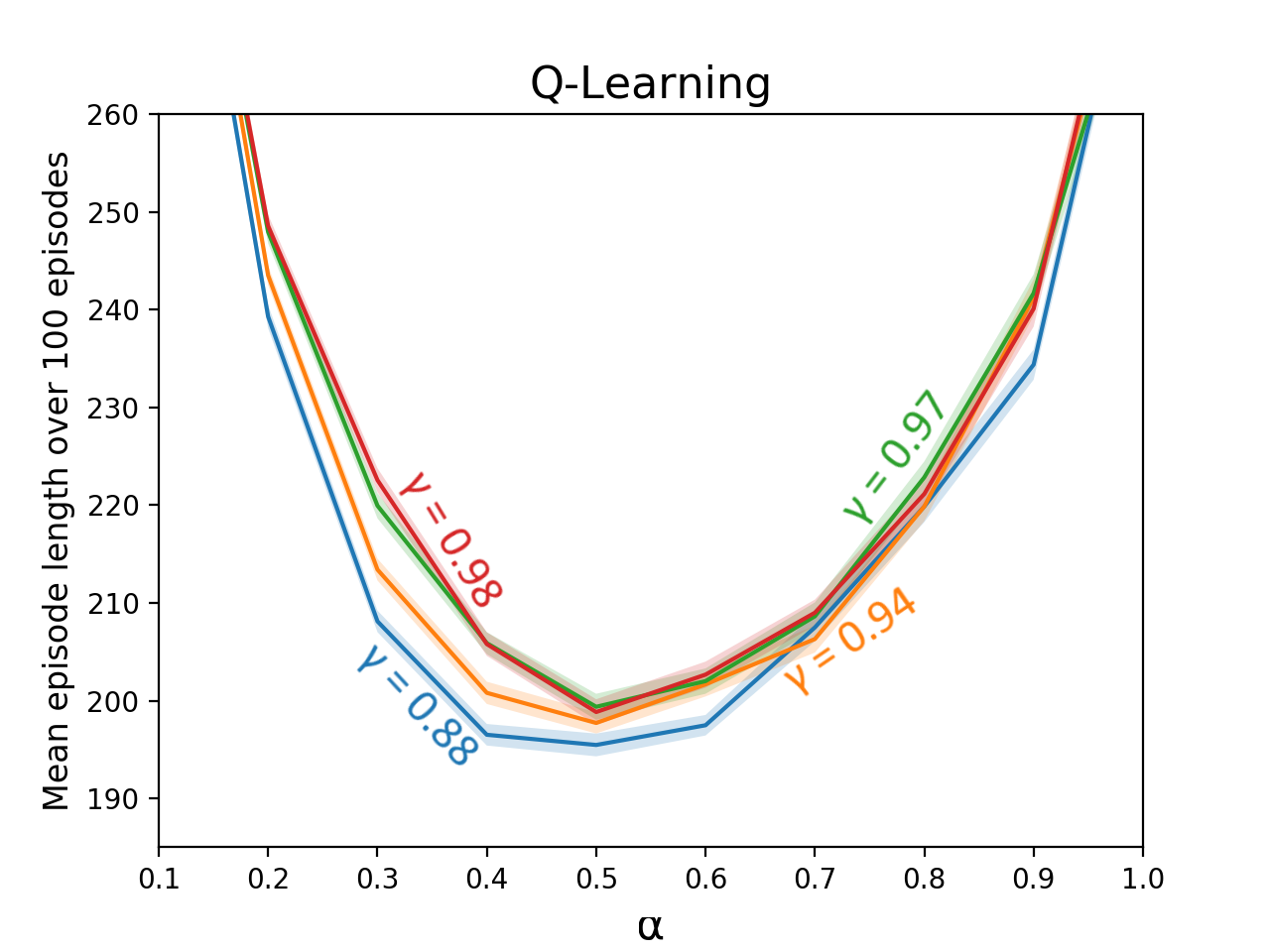}
	\caption[]{Mean episode lengths over 100 episodes of FHQ-learning and Q-learning with various step-sizes and horizons of interest. Results are averaged over 100 runs, and shaded regions represent one standard error.}
	\label{fig:results_slippery_maze}
\end{figure*}

In this section, we evaluate one-step FHQ-learning in a control problem. We hypothesize that when transitions are highly stochastic, predicting too far into the future results in unnecessarily large variance. Using fixed step sizes, we expect this to be an issue even in tabular settings. Both truncating the horizon and constant-valued discounting can address the variability of long term information, so we compare undiscounted FHQ-learning to discounted Q-learning.

We designed the \textit{slippery maze} environment, a maze-like grid world with 4-directional movement. The agent starts in the center, and hitting walls keep the agent in place. The ``slipperiness'' involves a $75\%$ chance that the agent's action is overridden by a random action. A reward of $-1$ is given at each step. The optimal deterministic path is 14 steps, but due to stochasticity, an optimal policy averages 61.77 steps.

Each agent behaved $\epsilon$-greedily with $\epsilon = 0.1$. We swept linearly spaced step-sizes, final horizons $H \in \{8, 16, 32, 48\}$ for FHQ-learning, and discount rates $\gamma \in \{0.875, 0.938, 0.969, 0.979\}$ for Q-learning. The discount rates were selected such that if $1 - \gamma$ represented a per-step termination probability of a stochastic process, the expected number of steps before termination matches the tested values of $H$. We performed 100 independent runs, and Figure \ref{fig:results_slippery_maze} shows the mean episode length over 100 episodes.

For FHQ-learning, it can be seen that if the final horizon is unreasonably short ($H = 8$), the agent performs poorly. However, $H = 16$ does considerably better than if it were to predict further into the future. With Q-learning, each discount rate performed relatively similar to one another, despite discount rates chosen to have expected sequence lengths comparable to the fixed horizons. This may be because they still include a portion of the highly variable information about further steps. For both algorithms, a shorter horizon was preferable over the full episodic return.

\subsection{Deep FHTD Control}
\label{subsec:deepFHTDcontrol}

We further expect FHTD methods to perform well in control with non-linear function approximation. FHTD's derivation assumes weights are separated by horizon. To see the effect of horizons sharing weights, we treated each horizon's values as linear neural network outputs over shared hidden layers. Use of this architecture along with parallelization emphasizes that the increased computation can be minimal. Due to bootstrapped targets being decoupled by horizon, we also expect deep FHTD methods to not need target networks.

In OpenAI Gym's \textit{LunarLander-v2} environment~\cite{gym2016}, we compared Deep FHQ-learning (DFHQ) with a final horizon $H = 64$ and DQN~\cite{dqn2015}. We restricted the neural network to have two hidden layers, and swept over hidden layer widths for each algorithm. We used $\gamma \in \{0.99, 1.0\}$, and behaviour was $\epsilon$-greedy with $\epsilon$ annealing linearly from $1.0$ to $0.1$ over 50,000 frames. RMSprop \cite{rmsprop2012} was used on sampled mini-batches from an experience replay buffer~\cite{dqn2015}, and ignoring that the target depends on the weights, DFHQ minimized the mean-squared-error across horizons:
\begin{align}
    \hat{G}_{t}^{h} &= R_{t+1} + \gamma\max_{a'}{Q^{h-1}(S_{t+1}, a'; \params)} \nonumber \\
    J(\params) &= \frac{1}{H}\sum_{h = 1}^{H}{\bigg(\hat{G}_{t}^{h} - Q^h(S_t, A_t; \params)\bigg)^2}
\end{align}

\begin{figure}[!tp]
    \centering
	\includegraphics[width=\figwidth]{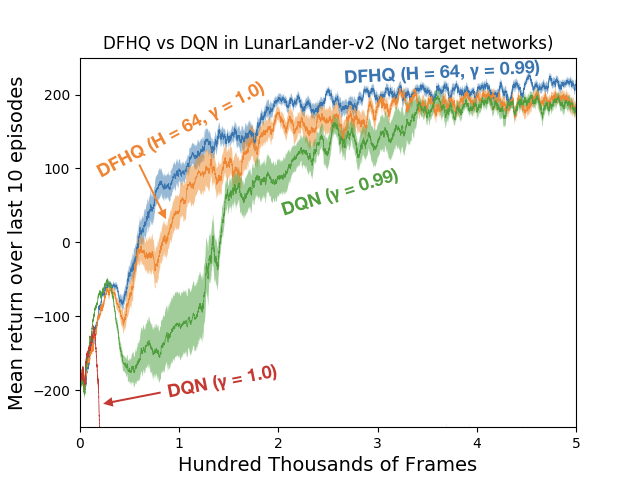}
	\caption[]{Mean return over last 10 episodes at each frame of DFHQ and DQN, without target networks, averaged over 30 runs. Shaded regions represent one standard error.}
	\label{fig:results_lunar_er}
\end{figure}

We performed 30 independent runs of 500,000 frames each (approximately 1000 episodes for each run). Figure \ref{fig:results_lunar_er} shows for each frame, the mean return over the last 10 episodes of each algorithm's best parameters (among those tested) in terms of area under the curve. Note that the results show DFHQ and DQN \textit{without} target networks. From additional experiments, we found that target networks slowed DFHQ's learning more than it could help over a run's duration. They marginally improve DQN's performance, but the area under the curve remained well below that of DFHQ.

It can be seen that DFHQ had considerably lower variance, and relatively steady improvement. Further, DFHQ was significantly less sensitive to $\gamma$, as DQN with $\gamma = 1$ immediately diverged. 
From the remainder of our sweep, DFHQ appeared relatively insensitive to large hidden layer widths beyond the setting shown, whereas DQN's performance considerably dropped if the width further increased.

DFHQ's good performance may be attributed to the representation learning benefit of predicting many outputs~\cite{unreal2016,hyperbolic2019}; in contrast with auxiliary tasks, however, these outputs are \textit{necessary tasks} for predicting the final horizon. An early assessment of the learned values can be found in Appendix \ref{appdx:deepvals}.

\section{Discussion and future work}

In this work, we investigated using \textit{fixed-horizon} returns 
in place of the conventional infinite-horizon return. We derived FHTD methods and compared them to their infinite-horizon counterparts in terms of prediction capability, complexity, and performance. We argued that 
FHTD agents are stable under function approximation and have additional predictive power. We showed that the added complexity can be substantially reduced via parallel updates, shared weights, and $n$-step bootstrapping. Theoretically, we proved convergence of FHTD methods with linear and general function approximation. Empirically, we showed that off-policy linear FHTD 
converges on a classic counterexample for off-policy linear TD. Further, in a tabular control problem, we showed that 
greedifying with respect to estimates of a short, fixed horizon could outperform doing so with respect to longer horizons. Lastly, we demonstrated that FHTD methods can scale well to and perform competitively on a deep reinforcement learning control problem.

There are many avenues for future work. Given that using shorter horizons may be preferable (Figure \ref{fig:results_slippery_maze}), it would be interesting if optimal weightings of horizons could be learned, rather than relying on 
the furthest horizon to act. Developing ways to handle the off-policyness of $n$-step FHTD control (See Appendix \ref{appdx:offpolicyness}), incorporating temporal abstraction, and experiments in more complex environments 
would improve our understanding of the scalability of our methods to extremely long horizon tasks. 
Finally, the applications to complex and hyperbolic discounting (Appendix \ref{appdx:arbitraryweights}), exploring the benefits of iteratively deepening the final horizon, and the use of fixed-horizon critics in actor-critic methods might be promising.

\subsubsection*{Acknowledgments}
The authors thank the Reinforcement Learning and Artificial Intelligence research group, Amii, and the Vector Institute for providing the environment to nurture and support this research. We gratefully acknowledge funding from Alberta Innovates -- Technology Futures, Google Deepmind, and from the Natural Sciences and Engineering Research Council of Canada.

\bibliographystyle{aaai}
\bibliography{references}

\newpage

\input{supplementary_fixedhorizon.tex}

\end{document}

%% file: supplementary_fixedhorizon.tex
\clearpage

\begin{appendices}

\section{Full Proof of Convergence for Linear Function Approximation}\label{appdx:convergence}
\subsection{Preliminaries}
We assume throughout a common probability space $(\Omega, \mathbb{P}, \Sigma)$. Our proof follows the general outline in~\citeauthor{melo2007q}~\shortcite{melo2007q}.

Assume we have a Markov decision process $(\mathcal{X}, \mathcal{A}, \mathcal{P}, r)$. $\mathcal{X} \subset \R^p$, the state-space, is assumed to be compact, the action space $\mathcal{A}$ is assumed to be finite (with $\sigma$-algebra $2^{\mathcal{A}}$), and $r : \mathcal{X} \times \mathcal{A} \times \mathcal{X} \to [r_{min}, r_{max}]$ is a bounded, deterministic function assigning a reward to every transition tuple. Let $\mathcal{F}$ by a $\sigma$-algebra defined on $\mathcal{X}$. The kernel $\mathcal{P}$ is assumed to be action-dependent, and is defined such that 
\begin{equation*}
    \mathbb{P}[X_{t + 1} \in U \mid X_t = x, A_t = a] = \mathcal{P}_a(x, U),
\end{equation*}

\noindent for all $U \in \mathcal{F}$. Throughout, we assume a fixed, measurable behaviour policy $\pi$ such that $\pi(a \mid x) > 0$ for all $a \in \mathcal{A}$ and $x \in \mathcal{X}$. With a fixed behaviour policy, we can define a new Markov chain $(X_t, A_t)$ first with a function $\mathcal{P}_\pi$:
\begin{equation*}
    \mathcal{P}_\pi((x, a), U \times A) := \sum_{b \in A} \int_U \pi(b \mid y) P_a(x, dy),
\end{equation*}

\noindent where $U \times A \subset \mathcal{F} \times 2^\mathcal{A}$. It is straightforward to see that $\mathcal{P}_\pi$ is a pre-measure on the algebra $\mathcal{F} \times 2^\mathcal{A}$ which can be extended to a measure on $\sigma(\mathcal{F} \times 2^\mathcal{A})$. 

\noindent To apply the results of ~\citeauthor{benveniste1990stochastic}~\shortcite{benveniste1990stochastic}, we must construct another Markov chain so that $H(\params_t, x_t)$ in ~\citeauthor[p.213]{benveniste1990stochastic}~\shortcite{benveniste1990stochastic} has access to the TD error at time $t$. In the interests of completeness, we provide the full details below, but the reader may safely skip to the next section.

We employ a variation of a standard approach, as in for example ~\citeauthor{tsitsiklis1999average}~\shortcite{tsitsiklis1999average}. Let us define a new process $M_t = (X_t, A_t, X_{t + 1}, A_{t + 1})$. The process $M_t$ has state space $\mathcal{M} := \mathcal{X} \times \mathcal{A} \times \mathcal{X} \times \mathcal{A}$ and $\sigma$-algebra $\sigma(\mathcal{F} \times 2^\mathcal{A} \times \mathcal{F}) 2^\mathcal{A}$, with kernel $\Pi$ defined first on $\mathcal{M} \times \mathcal{F} \times 2^\mathcal{A} \times \mathcal{F}$
\begin{align*}
    \Pi((x_t, a_t, x_{t + 1})&, U \times A \times V \times B) :=\\ 
    &\hspace{-1cm} 1_{(x_{t + 1}, a_{t + 1}}(U \times A) \mathcal{P}_\pi((x_{t + 1}, a_{t + 1}), V \times B).\nonumber
\end{align*}

\noindent Similar to before, it is straightforward to see that the above function is a pre-measure and can be extended to a measure on $\sigma(\mathcal{F} \times 2^\mathcal{A} \times \mathcal{F} \times 2^\mathcal{A})$ for each fixed $(x_t, a_t, x_{t + 1}, a_{t + 1})$. 

\begin{lemma}
$\Pi$ is a Markov kernel. 
\end{lemma}
\begin{proof}
It remains to show that $\Pi$ is measurable with respect to $(x_t, a_t, x_{t + 1}, a_{t + 1})$ for fixed $U \in \sigma(\mathcal{F} \times 2^\mathcal{A} \times \mathcal{F} \times 2^\mathcal{A})$.

We use Dynkin's $\pi-\lambda$ theorem. Define 
\begin{equation*}
 \mathcal{D} := \{U \in \sigma(\mathcal{F} \times 2^\mathcal{A} \times \mathcal{F}) : Q((\cdot, \cdot, \cdot, \cdot), U) \text{ is measurable}\}   
\end{equation*}
We have $\mathcal{F} \times 2^\mathcal{A} \times \mathcal{F} \times 2^\mathcal{A} \subset \mathcal{D}$ by construction of $\Pi$ above and the fact that $\mathcal{P}$ is a kernel. $\mathcal{F} \times 2^\mathcal{A} \times \mathcal{F} \times 2^\mathcal{A}$ is also a $\pi$-system (i.e., it is closed under finite intersections). $\mathcal{D}$ is a monotone class from using basic properties of measurable functions, so that $\mathcal{D} = \sigma(\mathcal{F} \times 2^\mathcal{A} \times \mathcal{F} \times 2^\mathcal{A})$ by the $\pi-\lambda$ theorem. Hence, $\Pi$ is a kernel. 
\end{proof}

\noindent The following is a convenient result. 

\begin{lemma}
Assume that $(X_t, A_t)$ is uniformly ergodic. Then $M_t = (X_t, A_t, X_{t + 1}, A_{t + 1})$ is also uniformly ergodic. 
\end{lemma}
\begin{proof}
From ~\citeauthor[p.389]{meyn2012markov}~\shortcite{meyn2012markov}, a Markov chain is uniformly ergodic iff it is $\nu_m$-small for some $m$. We will show that $M_t$ is $\eta_m$-small for some measure $\eta_m$. Since $(X_t, A_t)$ is uniformly ergodic, let $m > 0$ and $\nu_m$ a non-trivial measure on $\mathcal{F}$ such that for all $(x, a) \in \mathcal{X} \times \mathcal{A}$, $B \in \sigma(\mathcal{F} \times 2^\mathcal{A})$, 

\begin{equation}\label{eq:p-small}
    \mathcal{P}^m((x, a), B) \geq \nu_m(B).
\end{equation}

For any $x_{t + 1}, x_t \in \mathcal{X}, a_{t + 1}, a_t \in \mathcal{A}$, and $U \in \sigma((\mathcal{F} \times 2^\mathcal{A})^2)$, we can write 
\begin{align*}
    \Pi^{m + 1}&(x_t, a_t, x_{t + 1}, a_{t + 1}, U) =\\
    &\int_{(\mathcal{X} \times \mathcal{A})^{m + 1}} \mathcal{P}_\pi((x_{t + 1}, a_{t + 1}), dy_1) \mathcal{P}_\pi(y_{m + 1}, U^{y_{m + 1}})\\ \nonumber
    &\prod_{i = 1}^m \mathcal{P}_\pi(y_i, dy_{i + 1}), \nonumber
\end{align*}

\noindent where $U^y := \{x \in \mathcal{X} \times \mathcal{A} : (x, y) \in U\}$. Using \Cref{eq:p-small}, we have 

\begin{align*}
    \int_{(\mathcal{X} \times \mathcal{A})^{m + 1}} &\mathcal{P}_\pi((x_{t + 1}, a_{t + 1}), dy_1) \mathcal{P}_\pi(y_{m + 1}, U^{y_{m + 1}}) \\
    &\quad\prod_{i = 1}^m \mathcal{P}_\pi(y_i, dy_{i + 1})\\
    &\geq \int_{(\mathcal{X} \times \mathcal{A})^{m + 1}} \mathcal{P}_\pi((x_{t + 1}, a_{t + 1}), dy_1) \\
    &\quad\mathcal{P}_\pi(y_{m + 1}, U^{y_{m + 1}})\prod_{i = 1}^m \mathcal{P}(y_i, dy_{i + 1})\\
    &= \int_{\mathcal{X} \times \mathcal{A}} \mathcal{P}_\pi^m((x_{t + 1}, a_{t + 1}), dy) \mathcal{P}_\pi(y, U^y)\\
    &\geq \int_{\mathcal{X} \times \mathcal{A}} \nu_m(dy) \mathcal{P}_\pi(y, U^y).
\end{align*}

\noindent Let us define the bottom expression as a function $\eta_m(U)$. We claim that $\eta_m$ is a measure on $\sigma((\mathcal{F} \times 2^\mathcal{A})^2)$. First, $\eta_m(\varnothing) = 0$. Second, $(\cup_i C_i)^y = \cup_i C_i^y$ for $\{C_i\}_i$ disjoint, and such $\{C_i^y\}$ are themselves disjoint, otherwise if $x \in C_i^y \cap C_j^y$ for $i \neq j$, then $(x, y) \in C_i \cap C_j$, which is impossible by assumption of disjointedness of the $\{A_i\}$. Finally, $\mathcal{P}_\pi$ is itself a measure in the second argument. 

It remains to check that $\eta_m$ is not trivial. Set $U = (\mathcal{X} \times \mathcal{A})^2$. Then for all $y \in \mathcal{X} \times \mathcal{A}$, we have $U^y = \mathcal{X} \times \mathcal{A}$. Then 

\begin{equation*}
    \eta_m(U) = \int_{\mathcal{X} \times \mathcal{A}} \nu_m(dy) \mathcal{P}_\pi(y, \mathcal{X} \times \mathcal{A}) = \nu_m(\mathcal{X} \times \mathcal{A}) > 0.
\end{equation*}

\noindent The last inequality is by assumption of $\nu_m$ being non-trivial.
\end{proof}

Note that our $M_t$ corresponds to the $X_t$ in ~\citeauthor[p.213]{benveniste1990stochastic}~\shortcite{benveniste1990stochastic}. With this construction finished, we will assume in the following that whenever we refer to $\mathcal{X}$ or the Markov chain $X_t$, we are actually referring to $(\mathcal{X} \times \mathcal{A})^2$ or the Markov chain $M_t$ respectively. 

We assume that the step-sizes $\alpha_t$ of the algorithm satisfy the following:
$$
    \sum_{t} \alpha_t = \infty\quad\text{  and  }\quad
    \sum_{t} \alpha_t^2 < \infty.
$$


We write $\phi(x, a)$ for the feature vector corresponding to state $x$ and action $a$. We will sometimes write $\phi_x$ for $\phi(x, a)$. Assume furthermore that the features are linearly independent and bounded. 

We assume that we are learning $H$ horizons, and approximate the fixed-horizon $h$-th action-value function linearly.
\begin{equation*}
    Q^h_\params(x, a) :=  \params^{h, :} \phi(x, a),
\end{equation*}

\noindent where $\params \in \R^{H \times d}$. Colon indices denote array slicing, with the same conventions as in NumPy. For convenience, we also define $\params^{0, :} := 0$. 

We define the feature corresponding to the max action of a given horizon:
\begin{equation*}
    \phi_{\params}^*(x, h) := \phi(x, \argmax_{a} Q^h_\params(x, a)).
\end{equation*}

\noindent The update at time $t + 1$ for each horizon $h$ is given by 
\begin{align*}
    \params_{t + 1}^{h + 1, :} &:= \params_t^{h + 1, :} + \alpha_t (r(x, a, y) + \gamma \params_t^{h, :} \phi^*_{\params_t}(y, h) \\
    &\quad - \params_t^{h + 1, :} \phi_x)\phi_x^T.
\end{align*}
Here, $\gamma \in (0, 1]$.

\subsection{Results}
\setcounter{proposition}{0}
\begin{proposition}\label{prop:odex}
For $h = 1, ..., H$, the following ODE system has an equilibrium.
\begin{equation}\label{eq:odex}
    \dot{\params}^{h + 1, :} = \Ex\left[(r(x, a, y) + \gamma \params^{h, :} \phi^*_{\params}(y, h) - \params^{h + 1, :} \phi_x)\phi_x^T\right]
\end{equation}

\noindent Denote one such equilibrium by $\params_e$, and define $\tilde{\params} := \params - \params_e$. If, furthermore, we have that for $h = 1, ..., H$,
\begin{align}\label{eq:odex-bound}
    \gamma^2\Ex[(\params^{h, :} \phi^*_{\params}&(y, h) - \params_e^{h, :} \phi^*_{\params_e}(y, h))^2 ] <\\
    &\quad \Ex\left[(\params^{h + 1, :} \phi_x - \params_e^{h + 1, :} \phi_x)^2\right]\nonumber 
\end{align}
\noindent then $\params_e$ is a globally asymptotically stable equilibrium of \Cref{eq:odex}.

\end{proposition}

\begin{proof}
First, we show that there is at least one equilibrium of \Cref{eq:odex}. Finding an equilibrium point amounts to solving the following equations for all $h$:
\begin{equation*}
    \params^{h + 1, :} \Ex[\phi_x \phi_x^T] = \Ex[(r(x, a, y) + \gamma \params^{h, :} \phi^*_{\params_t}(y, h)) \phi_x^T]
\end{equation*}
Since we assume that the features are linearly independent, and using the fact that $\params^{0, :} = 0$, we can recursively solve these equations to find an equilibrium. 

Let $\params_e$ be the equilibrium point thus generated. Define $\tilde{\params} := \params - \params_e$ and substitute into \Cref{eq:odex} to obtain the following system. 


\begin{align}\label{eq:new-ode}
    \dot{\tilde{\params}}^{h + 1, :} &=  \Ex[(\gamma \params^{h, :} \phi^*_{\params}(y, h) - \gamma \params_e^{h, :} \phi^*_{\params_e}(y, h) \\
    &\quad - \tilde{\params}^{h + 1, :} \phi_x)\phi_x^T] \nonumber
\end{align}

\noindent The equilibrium $\tilde{\params} = 0$ of \Cref{eq:new-ode} corresponds to the equilibrium $\params = \params_e$ of \Cref{eq:odex}. By showing global asymptotic stability of 0 for \Cref{eq:new-ode} and using the change of variable $\tilde{\params} = \params - \params_e$, we will have global asymptotic stability of $\params_e$ for \Cref{eq:odex}.

Let us use the squared Euclidean norm $\|\cdot\|^2$ on $\R^{H \times d}$ as our Lyapunov function. Such a function is clearly positive-definite. Let $\tilde{\params}$ now denote a trajectory of \Cref{eq:new-ode}. Calculating,
\begin{align*}
    \dv{}{t}\|\tilde{\params}\|^2 &= \sum_{h = 0}^{H - 1}\dv{}{t} \left(\tilde{\params}^{h+1, :} \cdot \tilde{\params}^{h + 1, :}\right)\\
        &= 2\sum_{h = 0}^{H - 1} \dot{\tilde{\params}}^{h + 1, :} \cdot \tilde{\params}^{h + 1, :}\\
        &= 2 \sum_{h = 0}^{H - 1} \Ex[(\gamma \params^{h, :} \phi^*_{\params}(y, h) - \gamma \params_e^{h, :} \phi^*_{\params_e}(y, h) \\
        &\quad\quad - \tilde{\params}^{h + 1, :} \phi_x)\phi_x^T] (\tilde{\params}^{h + 1, :})^T\\
        &= 2\sum_{h = 0}^{H - 1} \Ex[(\gamma \params^{h, :} \phi^*_{\params}(y, h) - \gamma \params_e^{h, :} \phi^*_{\params_e}(y, h))\phi_x^T \\
    &\quad\quad (\tilde{\params}^{h + 1, :})^T] - \Ex\left[(\tilde{\params}^{h + 1, :} \phi_x)^2\right]\\
        &\leq 2\sum_{h = 0}^{H - 1} \sqrt{\gamma^2 \Ex\left[(\params^{h, :} \phi^*_{\params}(y, h) - \params_e^{h, :} \phi^*_{\params_e}(y, h))^2\right]}\\
        &\quad\quad \sqrt{\Ex\left[(\tilde{\params}^{h + 1, :} \phi_x )^2 \right]} - \Ex\left[(\tilde{\params}^{h + 1, :} \phi_x)^2\right]\\
        &< 0.
\end{align*}

\noindent We used H\"\o lder's inequality for the first inequality and \Cref{eq:odex-bound} for the last. The claim follows. 
\end{proof}

\noindent Defining $Q^*_{\params}(x, h) = \argmax_a Q_\params^h(x, a)$, we can rewrite \Cref{eq:odex-bound} as 
\begin{align}\label{eq:odex-bound-2}
    \Ex[(Q^{h + 1}(x, a) - &Q^{h + 1, e}(x, a))^2] >\\ &\gamma^2 \Ex\left[(Q^*_\params(x, h) - Q_{\params_e}^*(x, h))^2\right]. \nonumber
\end{align}
Effectively, \Cref{eq:odex-bound-2} means that the $h$-th fixed-horizon action-value function must be closer to the corresponding equilibrium, when taking the max action for each function, than the $(h + 1)$-th fixed-horizon action-value function is to its equilibrium when averaged across states and actions according to the behaviour policy. Intuitively, the functions for the previous horizons must have converged somewhat for the next horizon to converge. This condition can also be more easily satisfied by using a lower value of $\gamma$. 

\setcounter{theorem}{0}
\begin{theorem}
Viewing the right-hand side of the ODE system \Cref{eq:odex} as a single function $g(\params)$, assume that $g$ is locally Lipschitz in $\params$. 

Assuming \Cref{eq:odex-bound}, a fixed behaviour policy $\pi$ that results in a Markov chain $(\mathcal{X}, \Pi)$, and the assumptions in the Preliminaries, the iterates of FHQL converge with probability 1 to the equilibrium of the ODE system \Cref{eq:odex}.
\end{theorem}
\begin{proof}
We apply Theorem 17 of~\citeauthor[p.239]{benveniste1990stochastic}~\shortcite{benveniste1990stochastic}. Conditions (A.1)-(A.2) easily follow from the step-size assumption and from the existence of a transition kernel for our Markov chain, which we write here as $\Pi$ to keep to the notation in~\citeauthor{benveniste1990stochastic}~\shortcite{benveniste1990stochastic}. We also need to check (A.3)-(A.4). (A.3) and (A.4) (ii)-(iii) will be included into the verification of conditions (1.9.1)-(1.9.6) below, while we have that (A.4) (i) holds by assumption of $h$ being locally Lipschitz. 

\noindent It remains to verify the conditions (1.9.1) to (1.9.6). Let us write the invariant measure of $(\mathcal{X}, \Pi)$ as $\mu$. 

\paragraph{(1.9.1)}
The $H(\theta, x)$ (also written as $H_\theta(x)$) in~\citeauthor[p.239]{benveniste1990stochastic}~\shortcite{benveniste1990stochastic} corresponds in our case to the following matrix with components $i, j$ and with $\theta$ replaced by $\params$ : 

\begin{equation}\label{eq:H}
    [(r(x, a, y) + \gamma \params^{h, :} \phi_{\params}^*(y, h) - \params^{h + 1, :} \phi(x, a))\phi(x, a)^T]_j, 
\end{equation}

\noindent with $0 \leq h \leq H$, $1 \leq j \leq d$, where we suppress the $n$ index for clarity. Because we are assuming bounded features, and because $[(r(x, a, y) + \gamma \params^{h, j} \phi_{\params}^*(y, h) - \params^{h + 1, j} \phi(x, a))\phi(x, a)^T]_{h, j}$ depends only linearly on $\params$, the bound (1.9.1) is easily seen to be satisfied after, for example, expanding and applying the Cauchy-Schwarz inequality several times.

\paragraph{(1.9.2)}
The bound is trivially satisfied since in our case, $\rho_n := 0$. 

\paragraph{(1.9.3)}
This bound is satisfied since we assume that our state space is a compact subset of Euclidean space and is thus bounded. 

\paragraph{(1.9.4)}
We construct $\nu_\params$ explicitly, given the suggestion in ~\citeauthor[p.217]{benveniste1990stochastic}~\shortcite{benveniste1990stochastic}. Define 

\begin{equation*}
    \nu_\params(z) := \sum_{k \geq 0} \Pi^k(H_\params - \mu H_\params)(z).
\end{equation*}

\noindent We will show that the above series converges for all $z$. If it does, then it is straightforward to check that A.4 (ii) is satisfied. Since our chain is assumed to be uniformly ergodic, we have the existence of $K > 0$ and $\rho \in [0, 1)$ such that for all $x \in \mathcal{X}$, 

\begin{equation*}
    \sup_A |\Pi^n(x, A) - \mu(A)| < K \rho^n.
\end{equation*}

\noindent Note that $M$ depends implicitly upon $\pi$.

For any probability measures $P, Q$, we will use the following standard equality. 

\begin{align*}
    \frac{1}{2} \sup_{|f| \leq 1} & \left\{\int f(x) P(dx) - \int f(x) Q(dx)\right\} \\
    &= \sup_A |P(A) - Q(A)|.
\end{align*}

\noindent We let $\|H_\params(x)\|_\infty$ denote the uniform norm with respect to the argument $x$, which we will write as $\|H_\params\|_\infty$ for notational simplicity. Then,
\begin{align*}
    \nu_\params(z) &:= \sum_{k \geq 0} \Pi^k(H_\params - \mu H_\params)(z)\\
    &= (\|H_\params\|_\infty + 1) \sum_{k \geq 0} (\Pi^k - \Pi^k \mu) \frac{H_\params(z)}{\|H_\params\|_\infty + 1}\\
    &= (\|H_\params\|_\infty + 1) \sum_{k \geq 0} (\Pi^k - \mu) \frac{H_\params(z)}{\|H_\params\|_\infty + 1}\\
    &\leq  (\|H_\params\|_\infty + 1) \\
        &\quad \sum_{k \geq 0}\sup_{|f| \leq 1} \left\{\int f(x) \Pi^k(z, dx) - \int f(x) \mu(dx)\right \}\\
    &= 2 (\|H_\params\|_\infty + 1) \sum_{k \geq 0} \sup_A |\Pi^k(z, A) - \mu(A)|\\
    &\leq  2 (\|H_\params\|_\infty + 1) \sum_{k \geq 0}  K \rho^n < \infty.
\end{align*}

\noindent Given our assumption of linear function approximation, the functional form of $H_\params(x)$ in \Cref{eq:H} implies that $\|H_\params\|_\infty$ only grows linearly in $|\params|$, so we also have (1.9.4).

\paragraph{(1.9.5)}
This assumption is trivially satisfied because of our assumption of a fixed behaviour policy $\pi$, meaning that the stochastic kernel $\Pi$ and the function $\nu_\params$ do not depend on $\params$. 

\paragraph{(1.9.6)}
This condition is satisfied by our step-size assumptions with $\lambda = 1$.

~\\~\\ Finally, assumption b of Theorem 17 is satisfied by our \Cref{prop:odex}. The claim follows. 
\end{proof}

\section{Arbitrary Temporal Weighting of Return}\label{appdx:arbitraryweights}

Subtracting subsequent horizons' values allows for extracting the expected reward at each step, trivially allowing for arbitrarily re-weighting the rewards within a fixed horizon. This allows for weighting schemes beyond previous generalized discounting frameworks, such as allowing the discount factor to be 0 for a few steps, while still including future rewards. However, this approach is computationally expensive, particularly if this re-weighted return is to be maximized. In this section, we show the FHTD framework can still directly estimate returns under such weighting schemes.

While fixed-horizon returns don't rely on discounting for convergence, discounting allows for prioritizing near-term rewards within a fixed horizon, and induces a sense of urgency. To reiterate FHTD's TD targets:
\begin{equation*}
\hat{G}^h_t = R_{t+1} + \gamma V^{h-1}(S_{t+1})
\end{equation*}
Under complex-valued discounting~\cite{tddft2019}, this would allow for predicting the expected return's \textit{exact} discrete Fourier transform, because the transform assumes a fixed and known sequence length.
Knowing the sequence length also allows for incremental estimation of the \textit{fixed-horizon average reward}. This may have implications with non-linear function approximation, where outputs having similar magnitudes might be favored. This is characterized by the following TD targets:
\begin{equation}
\hat{G}^h_t = \frac{1}{h}R_{t+1} + \frac{h-1}{h} V^{h-1}(S_{t+1})
\label{eqn:avgfhtdtarget}
\end{equation}
Taking this further, if we only scale the sampled rewards (and not the bootstrapped value), the weight of each reward in the return is decoupled temporally. This allows for arbitrary temporal weighting schemes, including a notable alternative for exponential discounting up to a final horizon $H$:
\begin{equation}
\hat{G}^h_t = \gamma^{H - h}R_{t+1} + V^{h-1}(S_{t+1})
\label{eqn:rexpfhtdtarget}
\end{equation}
It can be seen that for $\gamma < 1$, many of the earlier horizons will have values that are close to zero, putting more reliance on later horizons being correct. The typical way of exponential discounting appears to put more burden on the earlier horizons being correct, as many of the later horizons will have similar values. This intuitively may have implications for trading off representational capacity by placing emphasis on certain horizons, but this was left for future work. Also of note, hyperbolic discounting~\cite{hyperbolic2019} up to a final horizon $H$ can use the targets:
\begin{equation}
\hat{G}^h_t = \frac{1}{1 + k(H - h)}R_{t+1} + V^{h-1}(S_{t+1})
\label{eqn:hyperfhtdtarget}
\end{equation}

\section{FHTD($\lambda$)}

Another way to perform multi-step TD learning is through TD($\lambda$) methods~\cite{rlbook2018}. These algorithms update toward the $\lambda$-return, a geometrically-weighted sum of $n$-step returns:
\begin{equation}
\hat{G}_t^\lambda = (1 - \lambda)\sum_{n=1}^{\infty}{\lambda^{n - 1}\hat{G}_{t:t+n}}
\label{eqn:deflambdareturn}
\end{equation}
It introduces a parameter $\lambda \in [0, 1]$ where $\lambda = 0$ gives one-step TD, and increasing $\lambda$ provides an efficient way to effectively include more sampled rewards into the estimate of the return. In the infinite-horizon setting, it makes computation depend on the size of the feature space, and no longer scales with the number of rewards to include in the estimate.
We can derive fixed-horizon TD($\lambda$), denoted FHTD($\lambda$), through this recursive form of its $\lambda$-return:
\begin{equation}
    \hat{G}^{\lambda, h}_t = R_{t+1} + \gamma\bigg((1 - \lambda)V^{h - 1}(S_{t+1}) + \lambda \hat{G}^{\lambda, h - 1}_{t + 1}\bigg)
    \label{eqn:deffhtdlambdareturn}
\end{equation}
Assuming the values are not changing, we can get the following sum of one-step FHTD TD errors:
\begin{align}
\delta^h_t &= R_{t+1} + \gamma V^{h - 1}(S_{t+1}) - V^h(S_{t})
\label{eqn:onestepfhtdtderror} \\
\hat{G}^{\lambda, h}_t &= V^h(S_t) + \sum^{H - 1}_{k = 0}{\delta^{h - k}_{t + k}\prod^{k}_{i = 1}{\gamma \lambda}}
\label{eqn:recfhtdlambdareturn}
\end{align}
This requires storage of feature vectors for the last $H$ states. Instead of storing rewards, it estimates the $\lambda$-return by backing up the current step's TD error, weighted by the product term in Equation \ref{eqn:recfhtdlambdareturn}. An observation is that FHTD($\lambda$) requires learning all $H$ value functions. Also, it checks the last $H$ states for each of the $H$ value functions, making its computation scale with $H^2$. Despite this, FHTD($\lambda$) allows for \textit{smooth} interpolation between one-step FHTD and fixed-horizon MC, and is convenient if one wants to dynamically vary the degree of bootstrapping.

\section{Additional Experiments}

\subsection{FHTD's Sample Complexity}\label{appdx:samplecomplexity}

We conjecture that FHTD's sample complexity is upper bounded by that of TD, assuming each horizon's initial values are identical to the initial values of TD. This is based on an observation that FHTD's updates to the final horizon's values will match TD's updates for the first $H - 1$ steps.

\begin{figure}[ht]
    \centering
	\includegraphics[width=0.8\linewidth]{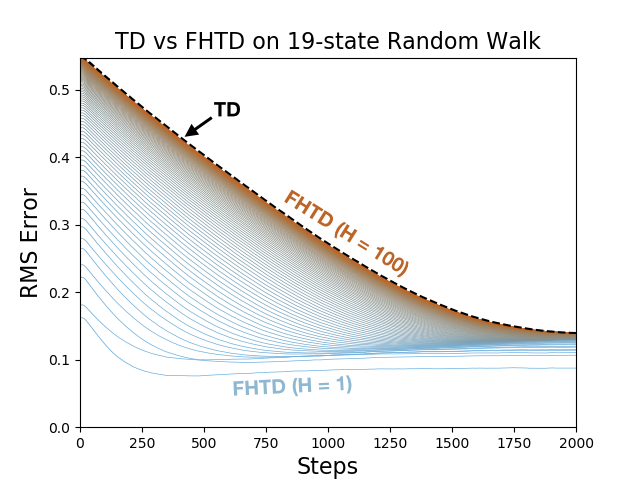}
	\caption[]{Prediction error after 2000 steps on the 19-state random walk. Results are averaged over 10000 runs, and standard errors are less than a line width.}
	\label{fig:results_19rw}
\end{figure}

We used a 19-state random walk, a tabular 1-dimensional environment where an agent starts in the center and randomly transitions to one of two neighboring states at each step. There is a terminal state on each end of the environment where transitioning to one of them gives a reward of $-1$, and transitioning to the other gives a reward of $1$. Treating it as an undiscounted, episodic task, the true values of FHTD's final horizon approach the true values of TD for sufficiently large $H$. In this experiment, we used a final horizon of $H = 100$, and a step size $\alpha = 0.5$. The true values for each horizon of FHTD, as well as the true values for TD, were computed with dynamic programming. After each step, we measured the root-mean-squared error between the learned and true values (uniformly averaged over states). We performed 10000 independent runs seeded such that each algorithm saw the same trajectories of experience, and Figure \ref{fig:results_19rw} shows the results after 2000 steps.

Evidently, short horizons plateau relatively quickly, with steady-state errors due to a fixed step size. As the horizon increases, FHTD's true values approach TD's true values, and FHTD's learning curves approach but don't cross that of TD. This supports our conjecture regarding FHTD's sample complexity, and suggests that errors propagate through horizons in a way that decomposes TD's error.

\subsection{Off-policyness of FHTD Control}
\label{appdx:offpolicyness}

Due to the inherent off-policyness of FHTD control, approximations are needed for the computational savings of $n$-step FHTD methods. We expect the off-policyness to primarily affect earlier horizons, as horizons will tend to agree on an action when looking sufficiently far into the future.

\begin{figure}[ht]
    \centering
	\includegraphics[width=0.8\linewidth]{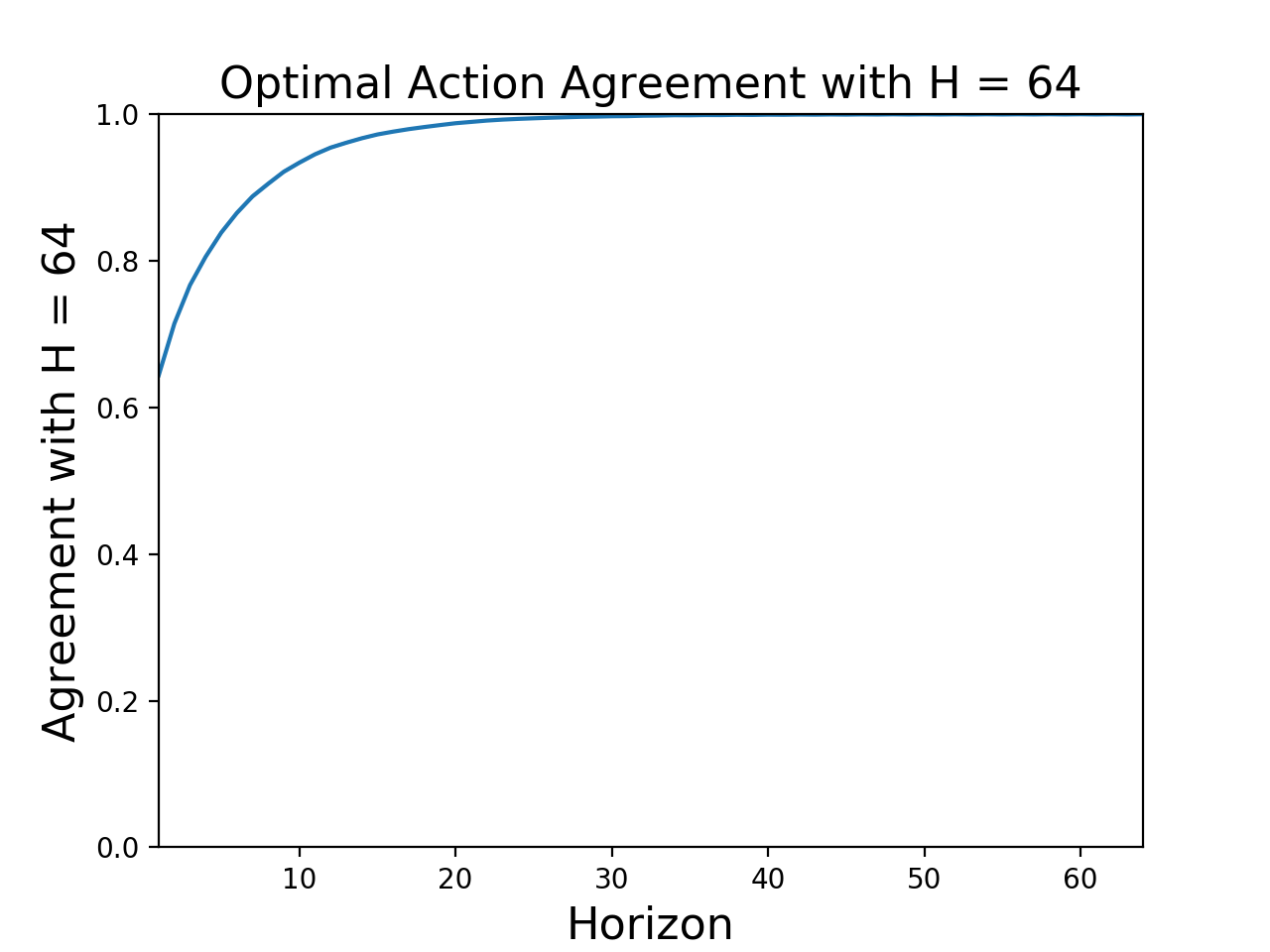}
	\caption[]{Optimal action agreement with the final horizon in random $8 \times 8$ grid worlds, uniformly averaged over states. Results are averaged over 1000 random grid worlds, and shaded regions represent one standard error.}
	\label{fig:results_offpolicyness}
\end{figure}

We randomly generated $8 \times 8$ grid worlds with 4-directional movement, where moving off of the grid keeps the agent in place. There were no terminal states, and upon environment initialization, the reward for transitioning into each state was a uniform random integer from the range $[-3, 3]$. We found that a final horizon of $H = 64$ was sufficient for identifying the optimal policy in these grid worlds, and used dynamic programming to compute each horizon's optimal policy. We computed how often each horizon's optimal action agreed with the optimal action for $H = 64$, uniformly averaged over states. Figure \ref{fig:results_offpolicyness} shows the results averaged over 1000 randomly generated $8 \times 8$ grid worlds.

The results support that most of the off-policyness occurs in the early horizons, and that later horizons tend to agree with the final horizon. A potential algorithm based on this observation is to learn the first few horizons with one-step FHTD, and then use $n$-step FHTD to skip horizons from there. 

\subsection{Visualizing Deep Value Estimates}
\label{appdx:deepvals}

\begin{figure}[!ht]
    \centering
	\includegraphics[width=0.8\linewidth]{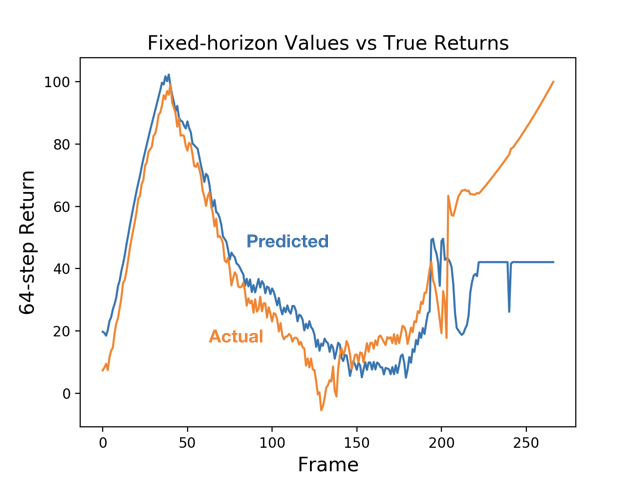}
	\caption[]{DFHQ's value estimates and the resulting (discounted) 64-step returns in a randomly sampled evaluation episode with frozen weights.}
	\label{fig:results_qs_dfhq}
\end{figure}

\begin{figure}[!ht]
    \centering
	\includegraphics[width=0.8\linewidth]{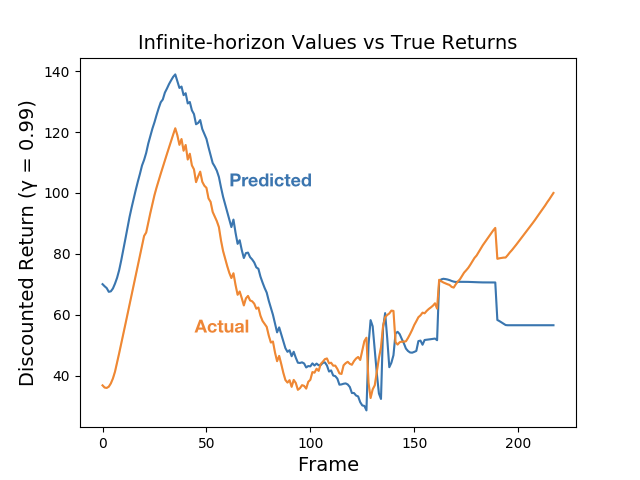}
	\caption[]{DQN's value estimates and the resulting discounted returns in a randomly sampled evaluation episode with frozen weights.}
	\label{fig:results_qs_dqn}
\end{figure}

Here we qualitatively compare DFHQ and DQN's value estimates with resulting returns. Each algorithm's weights were frozen after a run of 500,000 frames, and in evaluation episodes, informed an $\epsilon$-greedy policy with $\epsilon=0.05$. We logged the value estimate of each selected action, and computed the resulting returns from each state at the end of the episode. Figures \ref{fig:results_qs_dfhq} and \ref{fig:results_qs_dqn} show for each algorithm, how well the returns were predicted in a randomly sampled episode.

At a glance, DQN's estimates are consistent with Q-learning's well-known \textit{maximization bias}~\cite{rlbook2018}. DFHQ's estimates appear less impacted by this, and they matched the true returns relatively well. This may suggest why DFHQ performed better, and supports the possibility that it learns a better representation from predicting many outputs with shared hidden layers. However, we emphasize the distinction from work on auxiliary tasks~\cite{unreal2016,hyperbolic2019} because earlier horizons' values are necessary for constructing estimates for later horizons.

\subsection{Multi-step FHTD Policy Evaluation}

\afterpage{%
\begin{figure*}[!h]
    \centering
	\includegraphics[width=0.35\linewidth]{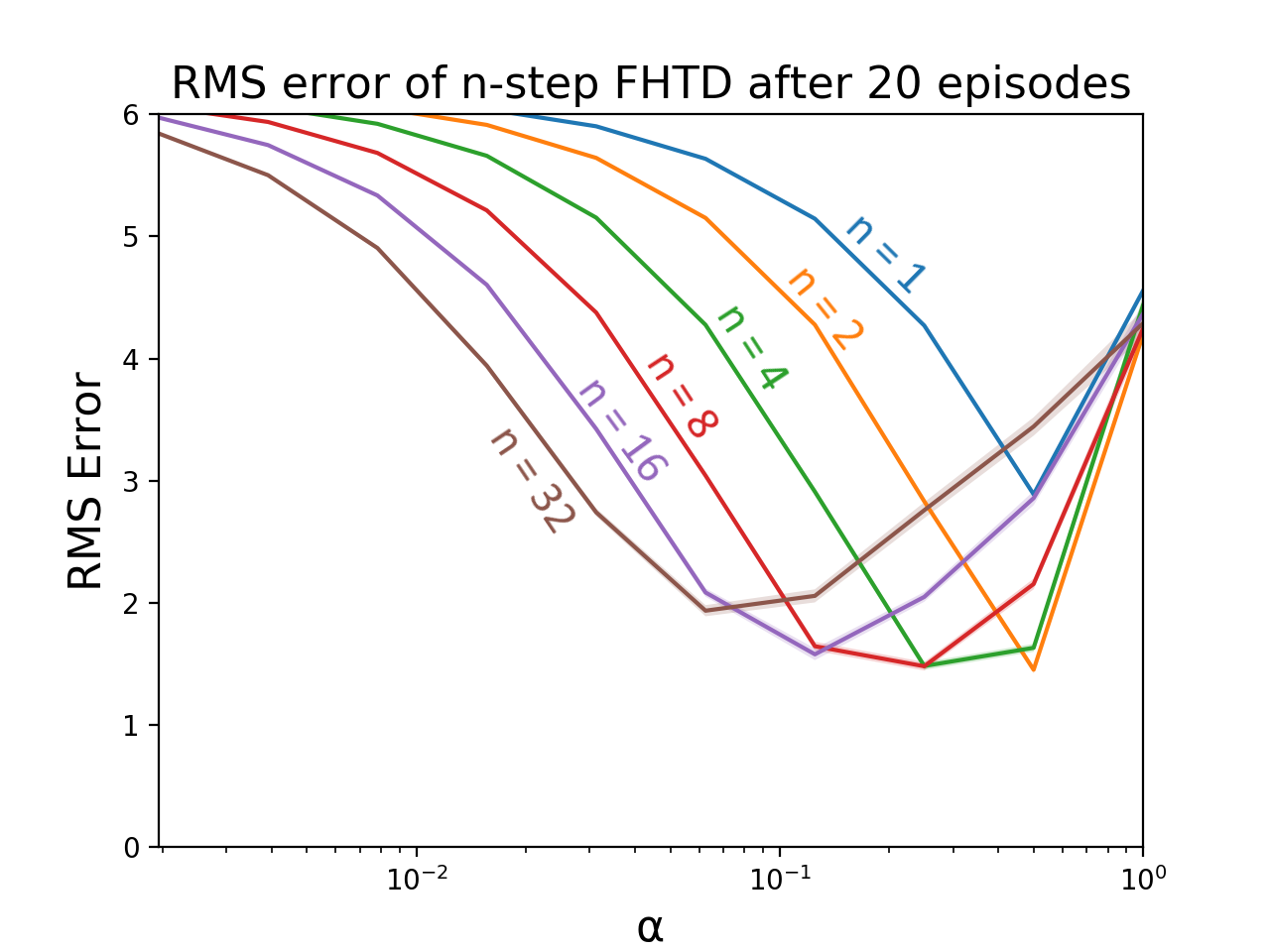}
	\includegraphics[width=0.35\linewidth]{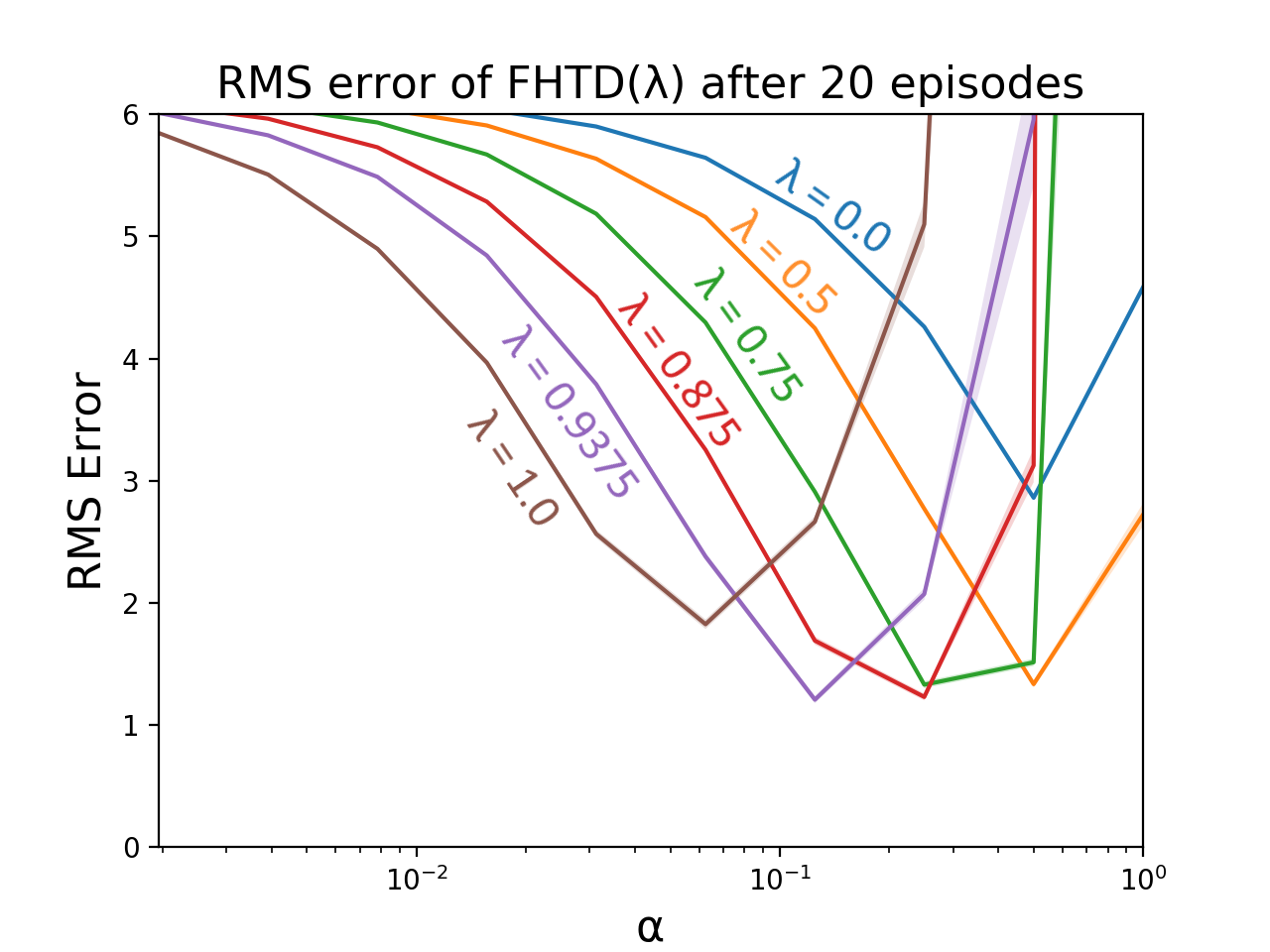}
	\includegraphics[width=0.35\linewidth]{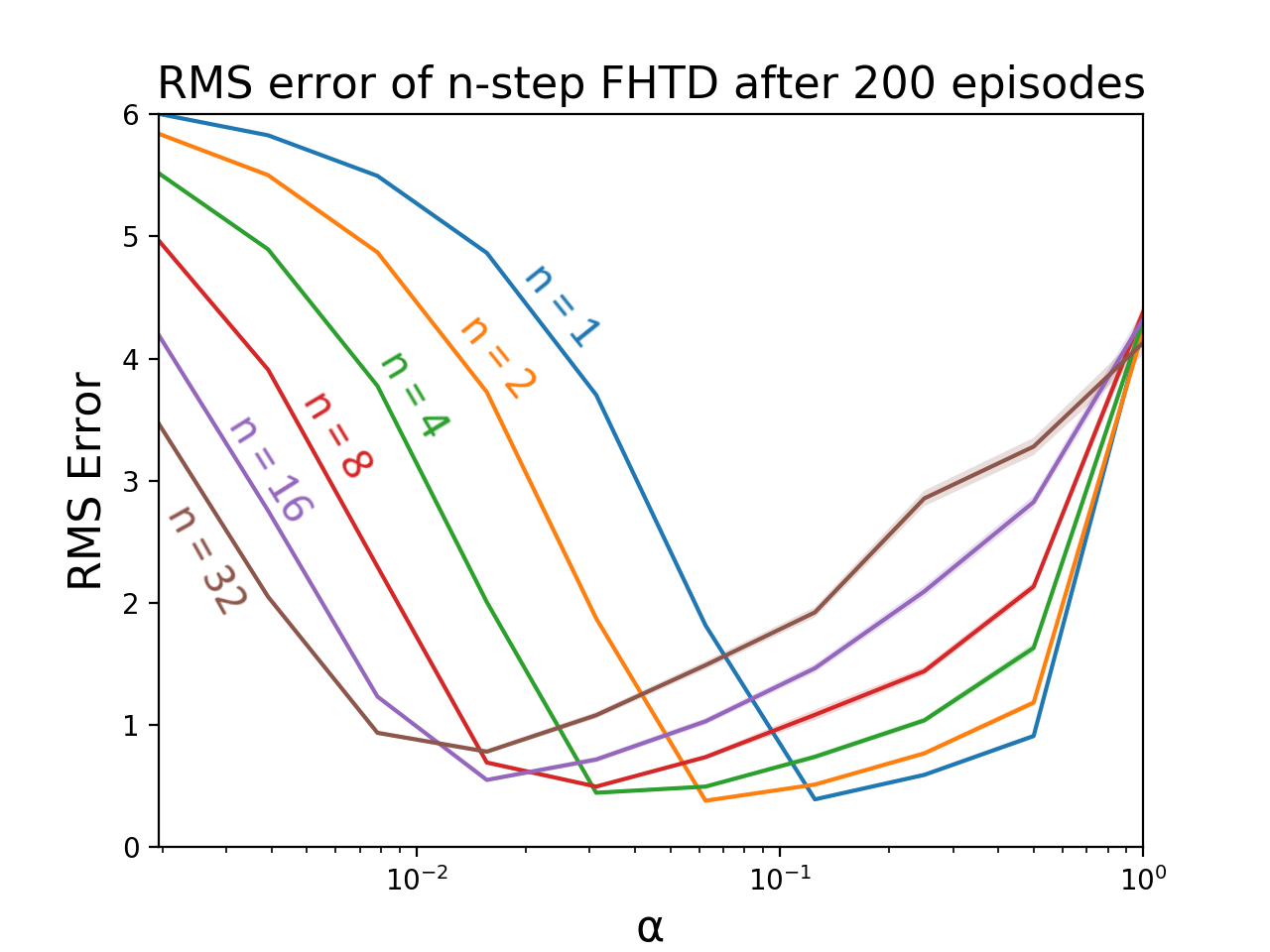}
	\includegraphics[width=0.35\linewidth]{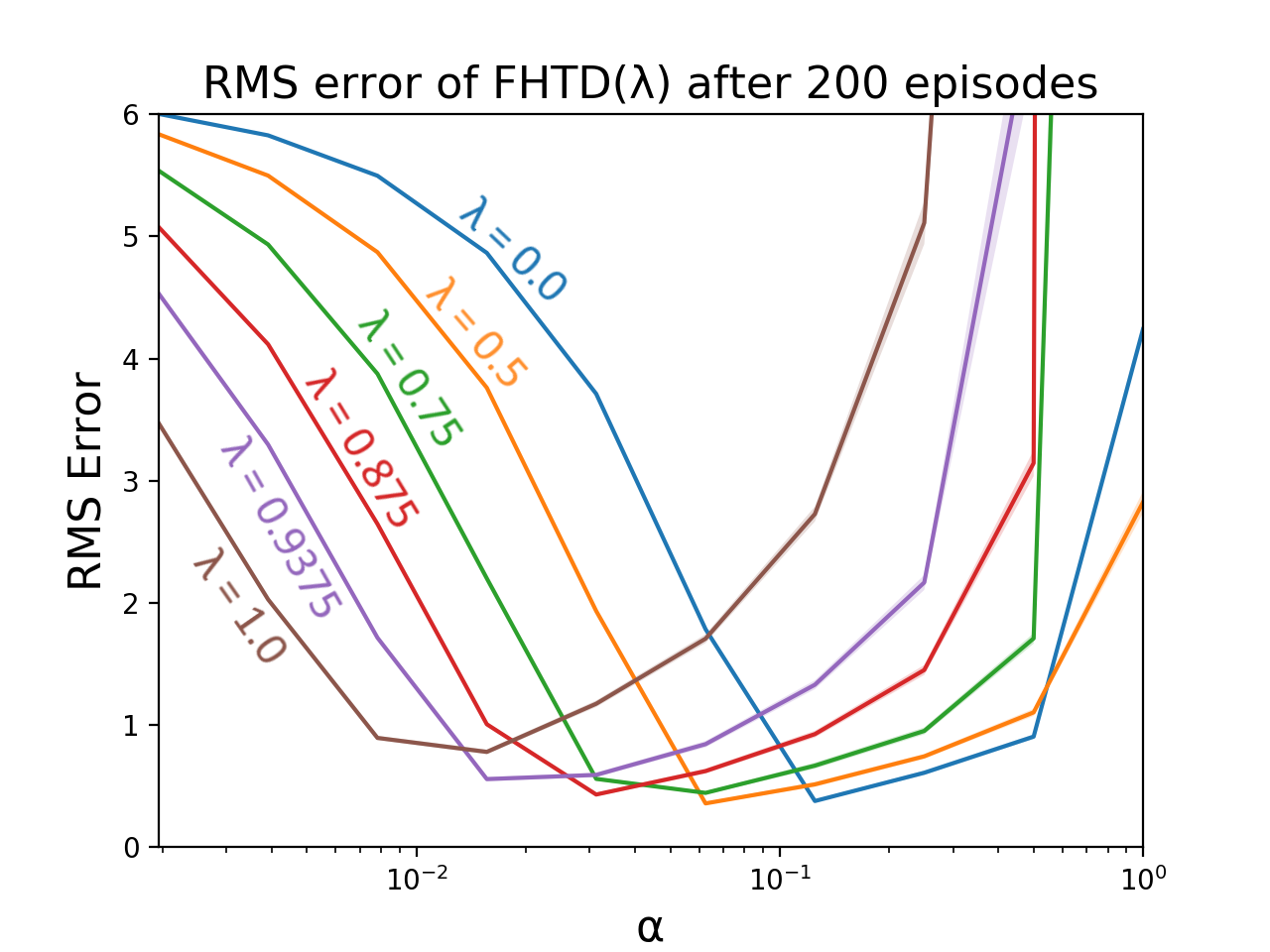}
	\caption[]{Prediction error after 20 and 200 episodes of multi-step FHTD algorithms with various step sizes and backup lengths on the checkered grid world. Results are averaged over 100 runs, and shaded regions represent one standard error. Note the log scale of the x-axis.}
	\label{fig:results_cgw}
\end{figure*}
\clearpage
}

In these experiments, we evaluate $n$-step FHTD and FHTD($\lambda$) in a policy evaluation task. We hypothesize that multi-step FHTD methods address a similar bias-variance trade-off to the infinite-horizon setting, such that intermediate values of $n$ and $\lambda$ can outperform either extreme. We also expect that $n$-step FHTD methods can tolerate larger step sizes relative to FHTD($\lambda$), as FHTD($\lambda$)'s derivation assumes that the value function is not changing.

We used the \textit{checkered grid world} environment~\cite{tddft2019}, as this environment's reward distribution emphasizes methods that can predict \textit{when} rewards occur. The environment is a 5 $\times$ 5 grid of states with terminal states on opposite corners. It has deterministic 4-directional movement, and moving into a wall keeps the agent in place. The agent starts in the center, and the board is colored with a checkered pattern which represents the reward distribution. One color represents a reward of $1$ upon entry, and a reward of $-1$ for the other. A reward of 11 is given at termination.

A final horizon of $H = 32$ was used, and each agent learned on-policy under an equiprobable random behavior policy. We swept over step sizes in negative powers of two, used $n \in \{1, 2, 4, 8, 16, 32\}$ for $n$-step FHTD, and $\lambda \in \{ 0.0, 0.5, 0.75, 0.875, 0.9375, 1.0\}$ for FHTD($\lambda$). The true value function for $H = 32$ was computed with dynamic programming, and after each episode, we measured the root-mean-square error in the 32nd horizon's values (uniformly averaged over states). We performed 100 independent runs, and Figure \ref{fig:results_cgw} shows the results after 20 and 200 episodes.

It can be seen that we get a similar trade-off where including more sampled rewards performs better early on, but one-step methods eventually catch up as its estimates become reliable. As expected, we also see that FHTD($\lambda$) is less stable for larger step sizes.

\section{Additional Experimental Details}

Below are the hyperparameter settings considered in our DFHQ and DQN results. Bolded values represent DFHQ's best parameter combination in terms of average episodic return over 500,000 frames (area under the curve). DQN's best parameter combination was identical apart from a marginal improvement with target networks.

\begin{center}
    \begin{tabular}{ |c|c| } 
     \hline
     \textbf{Parameter} & \textbf{Value(s)} \\
     \hline
     Per Episode Frame Limit & $5000$ \\
     \hline
     Replay Buffer Size & $10^{5}$ \\
     \hline
     Mini-batch Size & $32$ \\
     \hline
     RMSprop Learning Rate & $10^{-5}, \mathbf{10^{-4}}, 10^{-3}$ \\
     \hline
     Hidden Layer Widths & $128, \mathbf{256}, 512, 4096$ \\
     \hline
     DFHQ Final Horizon ($H$) & $32, \mathbf{64}$ \\
     \hline
     Discount Rate ($\gamma$) & $\mathbf{0.99}, 1.0$ \\
     \hline
     Target Net. Update Freq. & $\mathbf{1}, 100$ \\
     \hline
    \end{tabular}
\end{center}

\section{Environment Diagrams}

\phantom{Below are diagrams of the environments used.}

\begin{minipage}{\linewidth}
\begin{center}
	\includegraphics[width=0.5\linewidth]{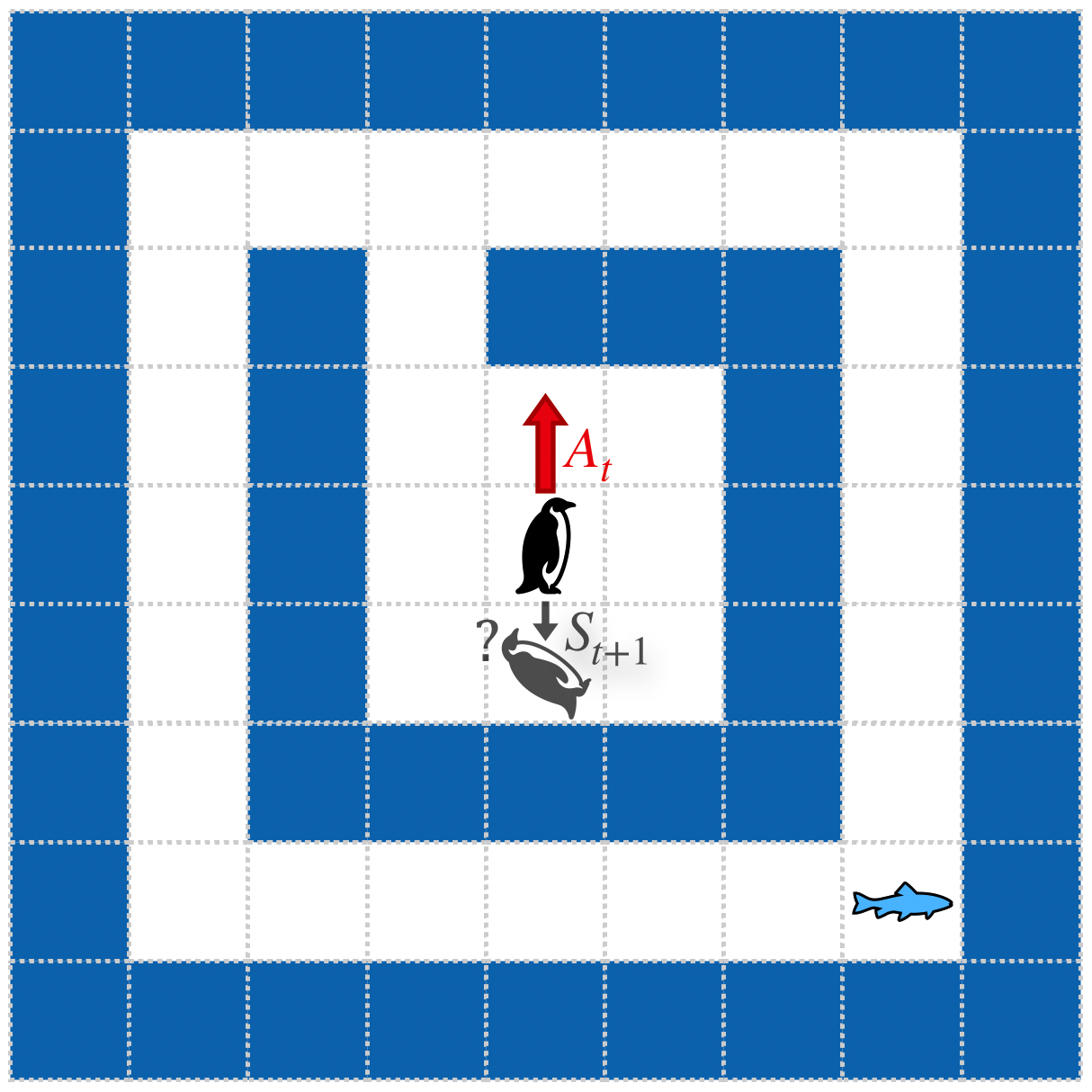}
	\captionof{figure}[Diagram of the slippery maze environment]{Diagram of the slippery maze environment. All rewards are $-1$ and reaching the bottom-right corner ends an episode. Each action selected has a $75\%$ chance of being overridden by a random action, unbeknownst to the agent.} 	\label{fig:env_slippery_maze}
    \vspace{\baselineskip}
\end{center}
\end{minipage}

\begin{minipage}{\linewidth}
\begin{center}
    \vspace{\baselineskip}
	\includegraphics[width=0.75\linewidth]{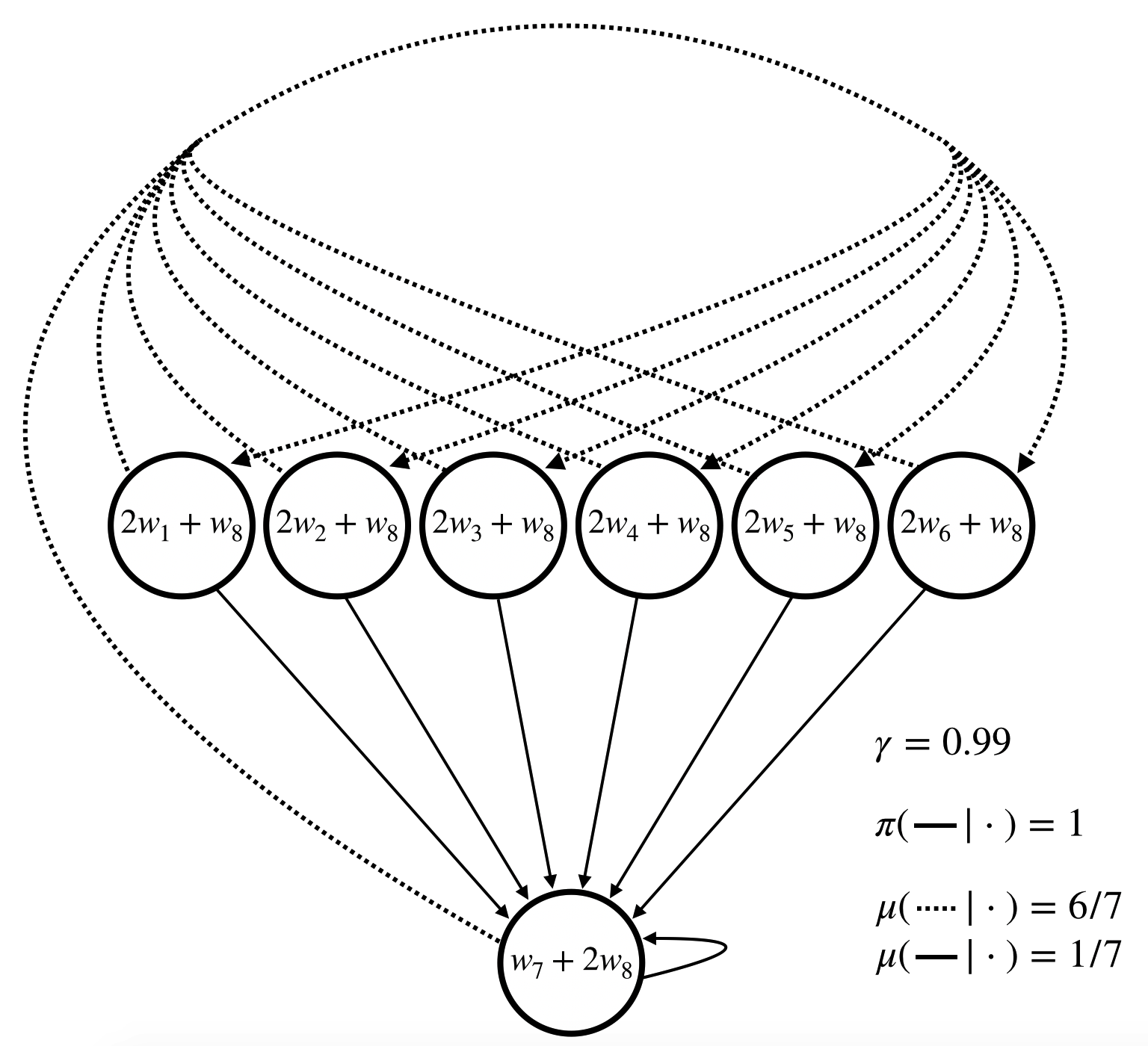}
	\captionof{figure}[Diagram of Baird's counterexample]{Diagram of Baird's counterexample. Each state's approximate state-value is shown by the linear expression inside each state. The objective is to predict the expected return from each state under a target policy which always chooses to transition to the 7th state.}\label{fig:env_baird}
    \vspace{\baselineskip}
\end{center}
\end{minipage}

\begin{minipage}{\linewidth}
\begin{center}
	\includegraphics[width=0.55\linewidth]{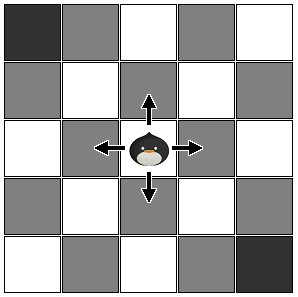}
	\captionof{figure}[Diagram of the checkered grid world environment]{Diagram of the checkered grid world environment. Transitioning into to a white square gives a reward of 1, transitioning into a gray square gives a reward of -1, and transitioning into the top-left or bottom-right corner ends an episode with a terminal reward of 11.}
	\label{fig:env_cgw}
\end{center}
\end{minipage}



\vfill

\section{Algorithm Pseudocode}



\begin{algorithm}[H]\small
\caption{\small Linear One-step FHTD for estimating $V^H \approx v^H_\pi$}
\label{alg:onestepfhtd}
\begin{algorithmic}
\State $\textbf{w} \leftarrow \textrm{Array of size } (H + 1) \times m$
\State $\textbf{w}_{[0]} \leftarrow [0 \textrm{ for } i \textrm{ in } \textrm{range}(m)]$
\State $s \sim p(s_0)$
\State $a \sim \pi(\cdot|s)$
\State $t \leftarrow 0$
\While{$t \neq t_{max}$}
    \State $s', r \sim p(s', r|s, a)$
    \For{$h = 1, 2, 3, ..., H$}
        \State $\delta \leftarrow r + \gamma \textbf{w}_{[h - 1]} \cdot \phi(s') - \textbf{w}_{[h]} \cdot \phi(s)$
        \State $\textbf{w}_{[h]} \leftarrow \textbf{w}_{[h]} + \alpha \delta \phi(s)$
    \EndFor
    \State $s \leftarrow s'$
    \State $a \sim \pi(\cdot|s)$
    \State $t \leftarrow t + 1$
\EndWhile
\end{algorithmic}
\end{algorithm}
\vspace{-\baselineskip}


\begin{algorithm}[H]\small
\caption{\small Linear One-step FHQ-Learning for estimating $Q^H \approx q^H_*$}
\label{alg:onestepfhqlearning}
\begin{algorithmic}
\State $\textbf{w} \leftarrow \textrm{Array of size } (H + 1) \times m$
\State $\textbf{w}_{[0]} \leftarrow [0 \textrm{ for } i \textrm{ in } \textrm{range}(m)]$
\State $s \sim p(s_0)$
\State $a \sim \mu(\cdot|s) \textrm{ (e.g. }\epsilon\textrm{-greedy w.r.t. }Q^H(s, \cdot))$
\State $t \leftarrow 0$
\While{$t \neq t_{max}$}
    \State $s', r \sim p(s', r|s, a)$
    \For{$h = 1, 2, 3, ..., H$}
        \State $\delta \leftarrow r + \gamma \max_{a'}{\big(\textbf{w}_{[h - 1]} \cdot \phi(s', a')\big)} - \textbf{w}_{[h]} \cdot \phi(s, a)$
        \State $\textbf{w}_{[h]} \leftarrow \textbf{w}_{[h]} + \alpha \delta \phi(s, a)$
    \EndFor
    \State $s \leftarrow s'$
    \State $a \sim \mu(\cdot|s)$
    \State $t \leftarrow t + 1$
\EndWhile
\end{algorithmic}
\end{algorithm}
\vspace{-\baselineskip}


\begin{algorithm}[H]\small
\caption{\small Linear $n$-step FHTD for estimating $V^H \approx v^H_\pi$}
\label{alg:nstepfhtd}
\begin{algorithmic}
\State $\textbf{w} \leftarrow \textrm{Array of size } (\frac{H}{n} + 1) \times m$
\State $\textbf{w}_{[0]} \leftarrow [0 \textrm{ for } i \textrm{ in } \textrm{range}(m)]$
\State $\bm{\Phi} \leftarrow \textrm{Array of size } n \times m$
\State $\textbf{R} \leftarrow \textrm{Array of size } n \times 1$
\State $s \sim p(s_0)$
\State $a \sim \pi(\cdot|s)$
\State $t \leftarrow 0$
\While{$t \neq t_{max}$}
    \State $\bm{\Phi}_{[t \ (\mathrm{mod}\ n)]} = \phi(s)$
    \State $s', r \sim p(s', r|s, a)$
    \State $\textbf{R}_{[t \ (\mathrm{mod}\ n)]} = r$
    \If{$t + 1 \geq n$}
        \State $r_{sum} \leftarrow \textrm{discountedsum}(\textbf{R})$
        \State $\phi_{old} \leftarrow \bm{\Phi}_{[(t + 1 - n) \ (\mathrm{mod}\ n)]}$
        \For{$h_{n} = 1, 2, 3, ..., \frac{H}{n}$}
            \State $\delta \leftarrow r_{sum} + \gamma^n \textbf{w}_{[h_{n} - 1]} \cdot \phi(s') - \textbf{w}_{[h_{n}]} \cdot \phi_{old}$
            \State $\textbf{w}_{[h_{n}]} \leftarrow \textbf{w}_{[h_{n}]} + \alpha \delta \phi_{old}$
        \EndFor
    \EndIf
    \State $s \leftarrow s'$
    \State $a \sim \pi(\cdot|s)$
    \State $t \leftarrow t + 1$
\EndWhile
\end{algorithmic}
\end{algorithm}


\begin{algorithm}[H]\small
\caption{\small Linear FHTD($\lambda$) for estimating $V^H \approx v^H_\pi$}
\label{alg:fhtdlambda}
\begin{algorithmic}
\State $\textbf{w} \leftarrow \textrm{Array of size } (H + 1) \times m$
\State $\textbf{w}_{[0]} \leftarrow [0 \textrm{ for } i \textrm{ in } \textrm{range}(m)]$
\State $\bm{\Phi} \leftarrow \textrm{Array of size } H \times m$
\State $s \sim p(s_0)$
\State $a \sim \pi(\cdot|s)$
\State $t \leftarrow 0$
\While{$t \neq t_{max}$} 
    \State $\bm{\Phi}_{[t \ (\mathrm{mod}\ H)]} \leftarrow \phi(s)$
    \State $s', r \sim p(s', r|s, a)$
    \For{$h = 1, 2, 3, ..., H$}
        \State $\delta^h = r + \gamma \textbf{w}_{[h-1]} \cdot \phi(s') -  \textbf{w}_{[h]} \cdot \phi(s)$
        \For{$i = 0, 1, 2, ..., H - h$}
            \State $\textbf{w}_{[h+i]} \leftarrow \textbf{w}_{[h+i]} + \alpha (\gamma\lambda)^i \delta^h \bm{\Phi}_{[(t - i) \ (\mathrm{mod}\ H)]}$
        \EndFor
    \EndFor
    \State $s \leftarrow s'$
    \State $a \sim \pi(\cdot | s)$
    \State $t \leftarrow t + 1$
\EndWhile

\end{algorithmic}
\end{algorithm}




\begin{algorithm}[H]\small
\caption{\small Linear FHQ($\lambda$) for estimating $Q^H \approx q^H_{*}$}
\label{alg:fhqlambda}
\begin{algorithmic}
\State $\textbf{w} \leftarrow \textrm{Array of size } (H + 1) \times m$
\State $\textbf{w}_{[0]} \leftarrow [0 \textrm{ for } i \textrm{ in } \textrm{range}(m)]$
\State $\bm{\Phi} \leftarrow \textrm{Array of size } H \times m$
\State $\Pi \leftarrow \textrm{Array of size } H \times H$
\State $s \sim p(s_0)$
\State $a \sim \mu(\cdot|s) \textrm{ (e.g. }\epsilon\textrm{-greedy w.r.t. }Q^H(s, \cdot))$ 
\State $t \leftarrow 0$
\While{$t \neq t_{max}$} 
    \State $\bm{\Phi}_{[t \ (\mathrm{mod}\ H)]} \leftarrow \phi(s, a)$
    \State $\Pi_{[t \ (\mathrm{mod}\ H)]} \leftarrow [\mathbbm{1}_{a = \argmax{Q^h(s,\cdot)}} \textrm{ for } h \in \{1, ..., H \}]$ 
    \State $s', r \sim p(s', r|s, a)$
    \For{$h = 1, 2, 3, ..., H$}
        \State $\delta^h = r + \gamma \max_{a'}{\big(\textbf{w}_{[h-1]} \cdot \phi(s', a')\big)} -  \textbf{w}_{[h]} \cdot \phi(s, a)$
        \State $e \leftarrow 1$
        \For{$i = 0, 1, 2, ..., H - h$}
            \State $\textbf{w}_{[h+i]} \leftarrow \textbf{w}_{[h+i]} + \alpha e \delta^h \bm{\Phi}_{[(t - i) \ (\mathrm{mod}\ H)]}$
            \If{$i \neq H - h$}
                \State $e \leftarrow e \gamma \lambda \Pi_{[(t - i) \ (\mathrm{mod}\ H), h + i - 1]}$
            \EndIf
        \EndFor
    \EndFor
    \State $s \leftarrow s'$
    \State $a \sim \mu(\cdot | s)$
    \State $t \leftarrow t + 1$
\EndWhile
\end{algorithmic}
\end{algorithm}





 
\clearpage

\end{appendices}
